%% file: main.tex
\documentclass[11pt]{article}

\usepackage{graphicx} 
\usepackage{amsmath}
\usepackage{amssymb}
\usepackage{amsthm}
\usepackage{tikz}
\usepackage{natbib}
\usepackage{authblk}

\usepackage[breaklinks, hidelinks]{hyperref}
\usepackage{url}

\usepackage{algorithm2e}
\DeclareMathOperator*{\argmin}{arg\,min}

\newtheorem{theorem}{Theorem}
\newtheorem{definition}{Definition}
\newtheorem{remark}{Remark}
\newtheorem{corollary}{Corollary}
\newtheorem{assumption}[theorem]{Assumption}
\newtheorem{assumptionalt}{Assumption}
\newtheorem{proposition}{Proposition}
\newtheorem{lemma}{Lemma}
\newtheorem{example}{Example}

\newenvironment{assumption_prime}[1]
  {\renewcommand{\theassumptionalt}{{#1}$'$}
   \begin{assumptionalt}}
  {\end{assumptionalt}}

\usetikzlibrary{arrows.meta}

\title{Boosting Causal Additive Models}
\date{\today}

\begin{document}
\author[1,2]{Maximilian Kertel}
\author[2,3]{Nadja Klein}
\affil[1]{ Technology Development Battery Cell,
       BMW Group,
       Munich, Germany}
\affil[2]{Department of Statistics,
       TU Dortmund University,
       Dortmund, Germany}
\affil[3]{Chair of Uncertainty Quantification and Statistical Learning, \newline
Research Center Trustworthy Data Science and Security (UA Ruhr)}
\title{Boosting Causal Additive Models}
\maketitle

\begin{abstract}
\input{00_abstract}
\end{abstract}

\section{Introduction}
\input{01_introduction}

\section{Causal Discovery}\label{sec:CausalDiscovery}
\input{02_causal_discovery}

\section{Background and Preliminaries}\label{sec:Background}\input{03_background}

\section{Boosting DAGs}\label{sec:BoostingDAGs} 
\input{04_BoostingDAGs/04_00_boosting_dags_intro}

\subsection{Assumptions}\label{sec:Assumptions}
\input{04_BoostingDAGs/04_01_assumptions}

\subsection{Main Theorem}\label{sec:main_theorem}
\input{04_BoostingDAGs/04_02_00boosting_dags}

\subsubsection{Boosting under Misspecification}\label{sec:boosting_misspec}
\input{04_BoostingDAGs/04_02_01_boosting_under_misspecification}

\subsubsection{Consistency of Variance Estimation}\label{sec:variance_est_consistent}
\input{04_BoostingDAGs/04_02_02boosting_error_estimates_are_consistent}

\subsection{Boosting DAGs for Large Dimensions}\label{sec:BoostingDAGs_high_dimensions}
\input{04_BoostingDAGs/04_03_boosting_dags_in_high_dimensions}

\section{Simulation Study}\label{sec:SimulationStudy}
\input{05_SimulationStudy}

\section{Conclusion}\label{sec:Conclusion}
\input{06_conclusion}

\newpage
\appendix

\section{Proof of Proposition~\ref{prop:consistency_empirical_score}}\label{sec:appendix_technical}
\input{Y_Appendix/A_proof_prop_1}
\section{Proof of Theorem~\ref{theorem:boosting_misspecification}}
\input{Y_Appendix/B_proof_theorem2}
\subsection{Upper-Bound of the RKHS Norm of Regression Function Estimate}\label{sec:appendix_upperbound_rkhs_norm}
\input{Y_Appendix/B1_Upperbound_h_norm}
\subsection{Results on Covering Numbers}\label{sec:appendix_covering}
\input{Y_Appendix/B2_covering_numbers}

\subsection{Results on Rademacher Complexities}\label{sec:appendix_rademacher}
\input{Y_Appendix/B3_rademacher_complexity}

\section{Proof of Theorem~\ref{theorem:consistency_variance}}\label{sec:appendix_fixed_design_convergence}
\input{Y_Appendix/C_fixed_design_convergence}

\section{Algorithm}\label{sec:appendix_algorithm}
\input{Y_Appendix/D_algorithm}

\bibliography{bibliography} 

\end{document}

%% file: 00_abstract.tex
We present a boosting-based method to learn additive Structural Equation Models (SEMs) from observational data, with a focus on the theoretical aspects of determining the causal order among variables.
We introduce a family of score functions based on arbitrary regression techniques, for which we establish necessary conditions to consistently favor the true causal ordering.
Our analysis reveals that boosting with early stopping meets these criteria and thus offers a consistent score function for causal orderings. 
To address the challenges posed by high-dimensional data sets, we adapt our approach through a component-wise gradient descent in the space of additive SEMs. Our simulation study underlines our theoretical results for lower dimensions and demonstrates that our high-dimensional adaptation is competitive with state-of-the-art methods. In addition, it exhibits robustness with respect to the choice of the hyperparameters making the procedure easy to tune.

%% file: 01_introduction.tex
Causal discovery is the process of deriving causal relationships between variables in a system. 
These help in improving decisions, predictions and interventions  \citep{improved_prediction_w_causal_discovery}. 
With the rapid growth of large-scale data sets in fields as healthcare, genetics \citep{aibar2017scenic}, or manufacturing \citep{kertel2022learning}, causal discovery has become increasingly important.
Traditionally, the researcher designs an experiment and intervenes on certain variables, that is, assigns for example a drug or placebo, which allows to estimate the causal effect of the manipulated variables.
Often however, it is less time, and effort consuming or more practical or ethical to collect data of a steady state of the system, where no variables are manipulated. For instance, in complex manufacturing domains, the number of input variables might be too large to conduct a design of experiments, and additionally those experiments would lead to a production downtime generating high costs. 

Thus, although observational data is often complex, noisy, or has confounding variables \citep{pmlr-v130-bhattacharya21a}, it is desirable to derive causal relationships from observational data rather than having to rely on impractical experimental studies.

In this work, we follow the assumption that the causal relationships between variables can be modeled as a Directed Acyclic Graph (DAG). This assumption implies that the impact between two variables flows in at most one direction, and there are no cyclic or self-reinforcing pathways. 
 It is the goal of the present work to identify the DAG from observational data. Existing algorithms for this task can be broadly classified into two categories. First,  constraint-based methods rely on statistical tests for conditional independence, but these approaches become computationally expensive and difficult apart from multivariate normal or multinomial distributions \citep{ZhangKernelTest, CondDepTestingIsHard}. 
Additionally, they assume faithfulness \citep{PetersBook}, and the identified graph is typically not unique \citep{SpirtesPC,kalischPC}. 

Instead, we focus on the second category for identifying DAGs from observational data, namely on Structural Equation Models (SEMs). SEMs  rely on the assumption that each variable is a function of other variables in the system and a perturbing noise term.
\cite{PetersAdditiveModelIdentifiable} show that if one restricts the functional relationships and the noise, called the Additive Noise Model (ANM), then the implied distribution is unique. Specifically, this is the case when the functional relationships are non-linear, and the noise term is Gaussian, as we assume throughout this work. This allows to identify the SEM from data.

Causal discovery using ANMs is an active field of research \citep{survey_causal_discovery}. 
Many recent works  leverage continuous acyclity characterisations \citep{lachapelle2019gradient, DAGGNN, DAGsNoTearsNonparametric, SAM,ng2022convergence, masked_dag_learning} and find the DAG using gradient-descent. However, opposed to earlier works \citep[as for example,][]{Lingam,CAM} for most machine learning methods the statistical behaviour is unknown \citep{kaiser2022unsuitability}.
In this context, a popular contribution is that of \cite{aibar2017scenic}, which successfully derives the cyclic graphical representation of a gene regulatory network using boosting. Our work is an extension of this approach towards acyclic graphs.

In this paper, we leverage the success of gradient-based methods towards statistical boosting for causal discovery in ANMs and investigate the underlying statistical behaviour. In particular, we show consistency and robustness  of our proposed method. 

Our main contributions are the following.
\begin{itemize}
    \item We propose a generic method for causal discovery based on unspecified regression techniques. We then state assumptions on the regression technique that lead to consistent causal discovery, see Proposition~\ref{prop:consistency_empirical_score}. 
    \item We show in Theorem~\ref{theorem:consistency_dag_boost} that $L^2$-boosting with early stopping fulfills the conditions of Proposition~\ref{prop:consistency_empirical_score}, that is, $L^2$-boosting can be employed for consistent causal discovery.
    \item We show in Theorem~\ref{theorem:boosting_misspecification} that $L^2$-boosting with early stopping avoids overfitting even under misspecification. Further, if the model is correctly specified, then $L^2$-boosting with early stopping is consistent, as we show in  Theorem~\ref{theorem:consistency_variance}. 
    \item We propose a variant of the boosting procedure for high-dimensional settings, when the number of variables $p$ is large.
    \item We conduct a simulation study demonstrating that our approach is competitive with state-of-the-art methods. We furthermore show that it is robust with respect to the choice of the hyperparameters and  thus easy to tune.
\end{itemize}
This paper is structured as follows. In Section~\ref{sec:CausalDiscovery} we discuss SEMs and required assumptions. Section \ref{sec:Background} reviews boosting and Reproducing Kernel Hilbert Spaces (RKHSs). Section~\ref{sec:BoostingDAGs} proposes the novel causal discovery method and shows its consistency. Section~\ref{sec:SimulationStudy} explores the results of Section~\ref{sec:BoostingDAGs} empirically and benchmarks different state-of-the-art algorithms with our approach, while Section~\ref{sec:Conclusion} concludes. Technical details and proofs can be found in the Appendix.

%% file: 02_causal_discovery.tex
Let $\mathbf{X} = \left(X_1, \ldots, X_p\right)^\top$ be a $p$-dimensional random vector. For any $S = \{s_1, \ldots, s_T\} \subset \{1, \ldots, p\}$ and a vector $\mathbf{x} \in \mathbb{R}^p$ we define $\mathbf{x}_S := (x_{s_1}, \ldots, x_{s_T})^\top$. 
Analogously, for the random vector $\mathbf{X}$ we set $\mathbf{X}_S := (X_{s_1}, \ldots, X_{s_T})^\top$. 
We order the elements in $S$ to make the representations unique, such that $s_1 < s_2 < \ldots < s_T$. For a Directed Acyclic Graph (DAG) $G$ on $X_1, \ldots, X_p$ we define the parents of $k \in \{1, \ldots, p\}$ denoted by $\mathbf{pa}_G(k)$ as those $j \in \{1, \ldots, p\}$ for which the edge $X_j \rightarrow X_k$ exists in $G$. We assume that there exists a DAG $G$ so that $\mathbf{X}$ follows a SEM with additive noise, that is
\begin{equation}\label{eq:SEMComplicated}
X_k = f_k(\mathbf{X}_{\mathbf{pa}_G(k)}) + \varepsilon_k.
\end{equation}
Here, $\varepsilon_k$ are i.i.d.~noise terms for $k = 1, \ldots, p$.

Every DAG has at least one topological ordering $\pi$ (that is, a permutation) on $\{1, \ldots, p\}$. We denote the nonempty set of topological orderings for $G$ by $\Pi(G)$. With $\pi \in \Pi(G)$ there can only be a directed path in $G$ from $X_j$ to $X_k$ if $\pi(j) < \pi(k)$ but not vice versa; see Figure~\ref{fig:example_sem} for an illustrating example. 

\begin{figure}[ht]
 \begin{minipage}[b][2.5cm][t]{0.49\linewidth}
   \centering
   \begin{align*}
     X_1 &= f_1(X_2) + \varepsilon_1 \\
     X_2 &= \varepsilon_2 \\
     X_3 &= f_3(X_2) + \varepsilon_3 \\
   \end{align*}
 \end{minipage}
 \begin{minipage}[b][2cm][t]{0.49\linewidth}
   \centering
\begin{tikzpicture}[->,shorten >=1pt,auto,node distance=2cm, thick,main node/.style={circle,draw,font=\sffamily\Large\bfseries}]

 \node[main node] (X2) {$X_2$}; \node[main node] (X3) [right of=X2] {$X_3$};\node[main node] (X1) [left of=X2] {$X_1$};

\path (X2) edge (X1); 
\path (X2) edge (X3); 

\end{tikzpicture} 
\end{minipage}
\caption{An example of a SEM on the left-hand side and its corresponding graph $G$ on the right-hand-side for $p=3$. The set of possible topological orderings is $\{(2,1,3), (2,3,1)\}$. For $\pi^0 = (2,3,1)$ it holds $X_1 = f_1(X_2) + \varepsilon_1 = f_{12}(X_2) + f_{13}(X_3) + \varepsilon_1$ with $f_{12} = f_1$ and $f_{13} = 0$.}
 \label{fig:example_sem}
\end{figure}

\subsection{Identifiability}
The goal of our analysis is to identify the graph $G$ from the distribution of $\mathbf{X}$. In general there is no one-to-one correspondence between the distribution $P(\mathbf{X})$ and the underlying SEM or $G$. 
However, if we impose appropriate restrictions on the noise terms $\{\varepsilon_k: k = 1, \ldots, p\}$ and the functions $\{f_k: k = 1, \ldots, p\}$, then the desired one-to-one correspondence between a distribution and a SEM exists \citep[see][Corollary 31]{PetersAdditiveModelIdentifiable}. In this case, we call the SEM identifiable. We consider SEMs with the following assumptions, which guarantee identifiability.
\begin{assumption}\label{ass:sem_type}
For the SEM of Equation~\eqref{eq:SEMComplicated} assume that $f_k$ has the additive decomposition 
$$f_k(\mathbf{x}_{\mathbf{pa}_G(k)}) = \sum_{j \in \mathbf{pa}_G(k)} f_{kj}(x_j), \quad k=1, \ldots, p,$$
where the $f_{kj}$ are three times differentiable, non-linear and non-constant for any $k=1, \ldots, p$ and $j \in \mathbf{pa}_G(k)$. Further, let $(\varepsilon_1, \ldots, \varepsilon_p)$ be a random vector of independent components, which are normally distributed with mean zero and standard deviations $\sigma_1, \ldots, \sigma_p > 0$. We call SEMs of this form Causal Additive Models \citep[CAMs;][]{CAM}.
\end{assumption}
Define $\varpi_\pi(k) = \{j : \pi(j) < \pi(k)\}$ as the predecessors of $k$ with respect to a permutation $\pi$. Thus for $p=3$ and $\pi=(2,1,3)$ it holds $\varpi_\pi(1) = \{2\}, \varpi_\pi(2) = \emptyset, \varpi_\pi(3) = \{1,2\}$.

Consider a CAM, which is characterized by functions $f_1, \ldots, f_p$, standard deviations $\sigma_1, \ldots, \sigma_p$, and a graph $G$ with topological ordering $\pi$. The implied density is
\begin{equation}\label{eq:density_decomposed}
p(\mathbf{x}) = \prod_{k=1}^p p\left(x_k | \mathbf{x}_{\mathbf{pa}_G (k)}\right) = \prod_{k=1}^p p\left(x_k | \mathbf{x}_{\varpi_\pi(k)})\right) .    
\end{equation}
Here, $p\left(x_k | \mathbf{x}_S\right)$ is the density of the conditional distribution of $X_k$ given $\mathbf{X}_S = \mathbf{x}_S$, which we assume to exist throughout this work. For $S = \emptyset$, we set $p\left(x_k | \mathbf{x}_S\right) = p\left(x_k\right)$.  
The second equality in Equation~\eqref{eq:density_decomposed} holds since $X_k$ is independent from its predecessors in $\pi$ given its parents \citep[see][Proposition 6.31]{PetersBook}, that is
\begin{equation}\label{eq:cond_independence}
X_k \perp X_{\varpi_\pi(k) \setminus \mathbf{pa}_G(k)} | \mathbf{X}_{\mathbf{pa}_G(k)}.    
\end{equation}
For any combination of $f_1, \ldots, f_p$, $G$ and any topological ordering $\pi$ of $G$, it trivially holds
$$
f_k = \sum_{j \in \mathbf{pa}_G(k)} f_{kj}(X_j) = \sum_{j \in \varpi_\pi(j)} \widetilde{f_{kj}}(X_j),
$$
where $\widetilde{f_{kj}} = f_{kj}$ if $j \in \mathbf{pa}_G(k)$ and $\widetilde{f_{kj}} = 0$ if $j \notin \mathbf{pa}_G(k)$. Consequently any CAM characterized by $f_1, \ldots, f_p,$ $G,$ $\sigma_1, \ldots, \sigma_p$ can be re-parameterized by $f_1, \ldots, f_p,$ 
 $\pi,$  $\sigma_1, \ldots, \sigma_p$, where $\pi$ is a topological ordering of $G$. Note that the latter parametrization is not unique, since $\pi$ can be chosen arbitrarily from the set of topological orderings of $G$. However, once the topological ordering $\pi$ is known,  $G$ can be found by identifying the parents of $X_k$ (those $j$ for which $f_{kj} \neq 0$) within $\varpi_{\pi}(k)$ for any $k=1, \ldots, p$. This is straightforward using pruning or feature selection methods \citep{Teyssier_causal_order, ShojaieKnownOrdering, CAM}. 
Thus, we simplify our objective and instead of searching for $G$, we aim to identify its topological ordering $\pi$. 

Consider a CAM characterized by $f_1, \ldots, f_p, G, \sigma_1, \ldots, \sigma_p$. It can be re-parameterized by the parameter tuple $\theta = \left(f_1, \ldots, f_p, \pi, \sigma_1, \ldots, \sigma_p\right)$.
 Using Equation~\eqref{eq:cond_independence} it follows that the implied conditional distribution of $X_k | \mathbf{X}_{\varpi_{\pi}(k)} = \mathbf{x}_{\varpi_{\pi}(k)}$ is Gaussian with mean $f_k\left(\mathbf{x}_{\varpi_{\pi}(k)}\right)$ and standard deviation $\sigma_k$. The implied density $p_\theta$ is given by
$$
\log(p_{\theta}(\mathbf{x})) = \sum_{k=1}^p \log\left(\frac{1}{\sigma_k} \phi\left( \frac{x_k - f_k (\mathbf{x}_{\varpi_{\pi}(k)})}{\sigma_k}\right)\right),
$$
where $\phi$ is the density function of a univariate standard normal distribution. From now on let $\mathbf{X}$ follow a CAM characterized by $\theta^0 = \left(f^0_1, \ldots, f^0_p, \pi^0, \sigma^0_1, \ldots, \sigma^0_p\right)$. To identify $\theta^0$ we define the population score function 
$$
\theta \mapsto \mathbb{E}_{p_{\theta^0}}\left[-\log\left(p_\theta(x)\right)\right].
$$
It holds 
$$
\mathbb{E}_{p_{\theta^0}}\left[-\log\left(p_{\theta^0}(x)\right)\right] \leq \mathbb{E}_{p_{\theta^0}}\left[-\log\left(p_\theta(x)\right)\right],
$$
and equality holds if and only if $p_{\theta^0} = p_{\theta}$ by the properties of the Kullback-Leibler divergence. For these minimal $\theta$, their ordering $\pi$ must be in $\Pi(G^0)$ by the identifiability.

\begin{figure}[ht]
\begin{minipage}[b][4.4cm][t]{0.49\linewidth}
    \centering
    \includegraphics[width=\linewidth]{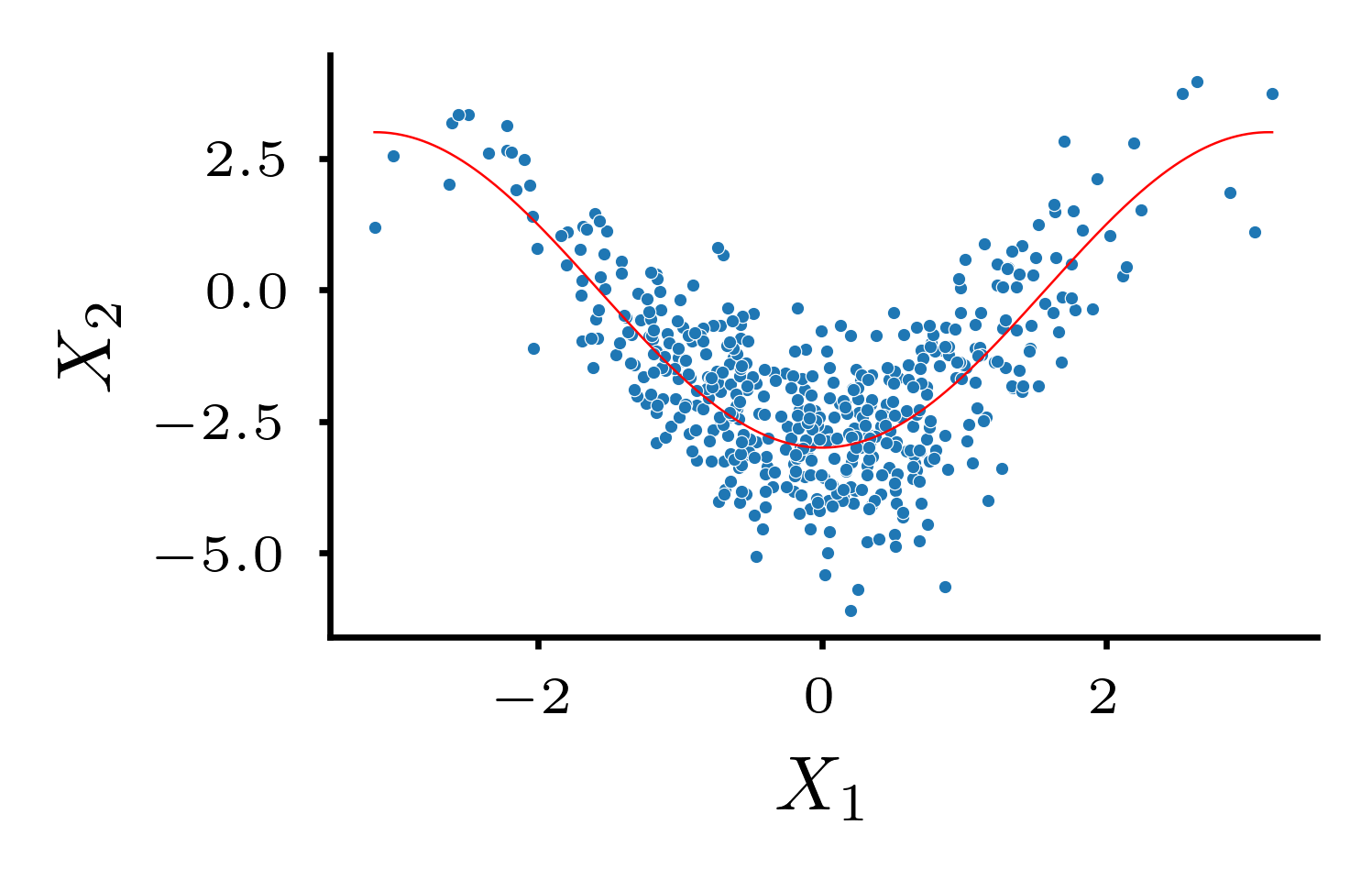}
\end{minipage}
\begin{minipage}[b][4.4cm][t]{0.49\linewidth}
    \centering
    \includegraphics[width=\linewidth]{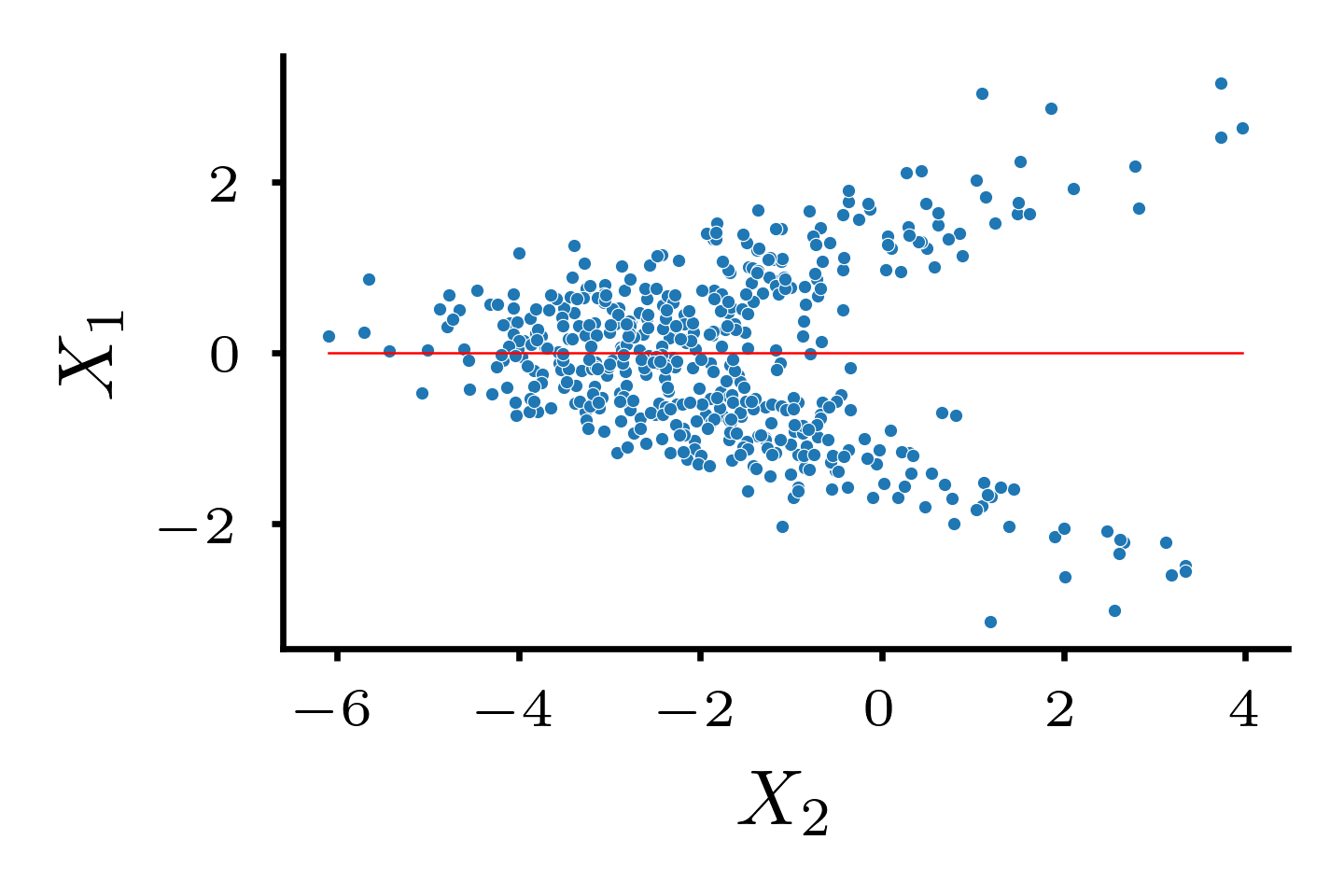}
\end{minipage}
\caption{The blue dots represent $500$ realizations of a distribution following a SEM with $p=2$ and $X_1 = \varepsilon_1 \sim \mathcal{N}(0, 1)$ and $X_2 = -3\cos(X_1) + \varepsilon_2$ with $\varepsilon_2 \sim \mathcal{N}(0, 1)$. On the left-hand-side we plot $X_2$ on the $y$-axis and $X_1$ on the $x$-axis, while on the right-hand-side it is vice versa. The red lines give the conditional mean functions. We see that $\argmin_{(f_1, f_2) \in \vartheta((1,2))} \sum_{k=1}^2\log(\sigma^2_{k, p_{\theta^0}, f_k, (1,2)})
= (0, -3\cos(x_1))$ and $ \argmin_{(f_1, f_2) \in \vartheta((2,1))} 
\sum_{k=1}^2\log(\sigma^2_{k, p_{\theta^0}, f_k, (2,1)}) = (0, 0)$. The distribution $X_1 - \mathbf{E}\left[X_1|X_2 = x_2\right]$ becomes bi-modal for larger values of $x_2$.
The unexplained noise (distance of blue dots to red line) is smaller on the left, which is the correct ordering, thus $S((1,2)) < S((2,1))$.}
\label{fig:two-dimensional-example}
\end{figure}

Let us consider the problem of minimizing the score function with respect to $\theta$.
Fixing $\pi$ and $f_1, \ldots, f_p$ in $\theta$ and minimzing with respect $\sigma_1, \ldots, \sigma_p$ leads to the minimizers
$$
\sigma_{k, p_{\theta^0}, f_k, \pi}^2 := \mathbb{E}_{p_{\theta^0}}\left[ \left(X_k - f_k (\mathbf{x}_{\varpi_{\pi}(k)})\right)^2\right] = \mathbb{E}_{p_{\theta^0}}\left[ \left(X_k - \sum_{j \in {\varpi_{\pi}(k)}} f_{kj}(X_{j}) \right)^2\right]
$$
for $k = 1, \ldots, p$. 
Hence, when minimizing the score function we only need to consider the subset of the parameter space $(f_1, \ldots, f_p,\pi, \sigma_1, \ldots,  \sigma_p)$, where $\sigma_k = \sigma_{k, p_{\theta^0}, f_k, \pi}, k=1, \ldots, p$, that is, the relevant parameter space reduces to $\left(f_1, \ldots, f_p, \pi\right)$.
Thus, it holds
\begin{align*}
\argmin_{\theta}\mathbb{E}_{p_{\theta^0}}\left[-\log\left(p_\theta(x)\right)\right] &= 
  \argmin_{\theta = (f_1, \ldots, f_p, \pi, \sigma_1 = \sigma_{1, p_{\theta^0}, f_1, \pi}, \ldots, \sigma_p = \sigma_{p, p_{\theta^0}, f_p, \pi})}\mathbb{E}_{p_{\theta^0}}\left[-\log\left(p_\theta(x)\right)\right] \\
  &= \argmin_{(f_1, \ldots, f_p, \pi)}\sum_{k=1}^p \log(\sigma^2_{k, {p_{\theta^0}}, f_k, \pi}) + C \\
  &= \argmin_{(f_1, \ldots, f_p, \pi)} \sum_{k=1}^p \log(\sigma^2_{k, {p_{\theta^0}}, f_k, \pi}),  
\end{align*}
$$
$$
with $C$ only depending on $p$. 
Denote the functions aligning with $\pi$ by 
\begin{align*}
\vartheta(\pi) = \left\{(f_1, \ldots, f_p) : f_k = \sum_{\pi(j) < \pi(k)} f_{kj}, f_{kj}: \mathbb{R} \rightarrow \mathbb{R}, f_{kj} \text{ is }\begin{array}{l}
\text{three times differentiable,} \\
\text{non-linear, and} \\
\text{non-constant.}
\end{array}\right\}.    
\end{align*}
We fix $\pi$ and optimize with respect to $f_1, \ldots, f_p$ to define a population score on the orderings
\begin{equation}\label{eq:population_score_function}
S(\pi) = \min_{(f_1, \ldots, f_p) \in \vartheta(\pi)} \sum_{k=1}^p\log(\sigma^2_{k, p_{\theta^0}, f_k, \pi}).
\end{equation}
By identifiability, $S(\pi)$ is minimal if and only if $\pi^0 \in \Pi(G^0)$, that is 
\begin{equation}\label{eq:population_score_chooses_correctly}
\sum_{k=1}^p\log\left(\left(\sigma_k^0\right)^2\right) = S(\pi^0) < S(\pi) \, \forall \pi^0 \in \Pi(G^0), \pi \notin \Pi(G^0).    
\end{equation}
Intuitively, the score $S(\pi)$ measures how much variance remains when any $X_k$ is regressed on using its predecessors $\mathbf{X}_{\varpi_\pi(k)}$. 
An example for $p=2$ is depicted in Figure~\ref{fig:two-dimensional-example}.

\subsection{Estimation of the Ordering}
In practice we are unaware of the true parameter tuple $\theta^0$ but observe $N$ realizations $\mathbf{x}^N := \left(\mathbf{x}_1, \ldots, \mathbf{x}_N\right)$ of $\mathbf{X}$ with density $p_{\theta^0}$, where $\mathbf{x}_\ell \in \mathbb{R}^p, \ell = 1, \ldots, N$ and $\mathbf{x}_\ell = \left(x_{\ell 1}, \ldots, x_{\ell p}\right)^\top$.
It is natural to propose the empirical version of the population score function (\ref{eq:population_score_function})
\begin{equation}\label{eq:score_function_permutation}
\widehat{S}(\pi) = \sum_{k=1}^p \log(\widehat{\sigma}^2_{k, \widehat{f}_{k, \pi}}),    
\end{equation}
where
$$
\widehat{\sigma}^2_{k, \widehat{f}_{k, \pi}} = \frac{1}{N} \sum_{\ell = 1}^N \left(x_{\ell k} -  \widehat{f}_{k, \pi}(\mathbf{x}_{\ell \varpi_\pi(k)})\right)^2 = \frac{1}{N} \sum_{\ell = 1}^N \left(x_{\ell k} -  \sum_{j \in \varpi_\pi(k)}\widehat{f}_{kj, \pi}(x_{\ell j})\right)^2.
$$
Here $\widehat{f}_{k, \pi} = \sum_{j \in \varpi_\pi(k)}\widehat{f}_{kj, \pi}$ is a regression function estimate using data $(\mathbf{x}_{\ell \varpi_\pi(k)}, x_{\ell k}), \ell = 1, \ldots, N$ with the convention $\mathbf{x}_{\ell S} := \mathbf{x}_{{\ell}_{S}} = 
\left(\mathbf{x}_{\ell s_1}, \ldots, \mathbf{x}_{\ell s_T}\right)$ for $S = \{s_1, \ldots, s_T\}$ introduced before. Although the regression estimates depend on the data and thus $N$, we omit the additional index for better readability.
\begin{remark}
In contrast to the population version \eqref{eq:population_score_function}, it is unclear whether \eqref{eq:score_function_permutation} is also minimized at $\pi^0 \in \Pi(G^0)$ even for $N \rightarrow \infty$. This is due to the fact, that the regression functions $\widehat{f}_{k, \pi}$ and the prediction errors $\widehat{\sigma}^2_{k, \widehat{f}_{k, \pi}}$
are estimated from a finite sample.
\end{remark}
Since any regression function estimator leads to a score function $\widehat{S}$, the following remark gives some intuition on necessary properties through two ``extreme'' examples.
\begin{remark}\label{remark:intuition_prop_1}
    \begin{enumerate}
        \item Let the regression estimator interpolate the data, that is, $\widehat{f}_{k, \pi}(x_{\ell, \varpi_\pi(k)}) =x_{\ell k}$ for all $\ell = 1, \ldots, N$, $k=1, \ldots, p$ and all $\pi$. Then the regression estimator is overfitting. In that case any permutation $\pi$ has a score diverging to $-\infty$. 
        \item Let $\widehat{f}_{k, \pi}(x_{\ell, \varpi_\pi(k)}) =0$ for all $\ell = 1, \ldots, N$, $k=1, \ldots, p$ and all $\pi$. Then the regression estimator is underfitting. Again, any ordering $\pi$ has the same score, which is $\sum_{k=1}^p \log\left(\frac{1}{N}\sum_{\ell=1}^N x_{\ell k}^2\right)$. 
    \end{enumerate}
    In both cases, $\widehat{S}$ cannot identify an optimal ordering $\pi^0 \in \Pi(G^0)$ even with infinite data. Intuitively, we need a regression estimator, that 
    \begin{enumerate}
        \item is not overfitting and preserves the non-explainable noise, and
        \item provides estimates that are close to the true regression functions $f_1^0, \ldots, f_p^0$.
    \end{enumerate}
\end{remark}
  In this work, we apply $L^2$-boosting regression in conjunction with early stopping \newline \citep{buehlmannYu, raskutti2014earlystopping} and show that the resulting score of Equation~\eqref{eq:score_function_permutation} prefers consistently a $\pi^0 \in \Pi(G^0)$.
Proposition~\ref{prop:consistency_empirical_score} below formalizes the intuition of Remark~\ref{remark:intuition_prop_1} and states necessary conditions on the regression function estimator. 
In the following Definition~\ref{def:overfitting}, $Y$ takes the role of $X_k$, while $\widetilde{\mathbf{X}}$ takes the role of $\mathbf{X}_{\varpi_\pi(k)}$ .
\begin{definition}[Non-overfitting]\label{def:overfitting}
    Let $\widehat{f}$ be a regression estimate based on $N$ i.i.d.~samples \newline $\left(\widetilde{\mathbf{x}}_1, y_1\right), \ldots, \left(\widetilde{\mathbf{x}}_N, y_N\right)$ from $(\widetilde{\mathbf{X}}, Y)$. We say the estimator is not overfitting w.r.t. $(\widetilde{\mathbf{X}}, Y)$, if 
    $$
    \left|\frac{1}{N}\sum_{\ell = 1}^N \left(y_\ell - \widehat{f}(\widetilde{\mathbf{x}_\ell})\right)^2 - \mathbb{E}_{\widetilde{\mathbf{X}}, Y}\left[\left(Y - \widehat{f}(\widetilde{\mathbf{X}})\right)^2 \right]\right|
    $$
    converges to $0$ in probability for $N \rightarrow \infty$.
\end{definition}

\begin{proposition}\label{prop:consistency_empirical_score}
    Let  Assumption~\ref{ass:sem_type} hold. Then, if the regression estimator is such that
    \begin{enumerate}
        \item $\widehat{f}_{k, S}$ is not overfitting with respect to $(\mathbf{X}_S, X_k)$ for any combination of $k = 1, \ldots, p$ and $S \subset\{1, \ldots, p\} \setminus \{k\}$ according to Definition~\ref{def:overfitting} and
        \item $\widehat{\sigma}^2_{k, \widehat{f}_{k, \pi^0}} \overset{\mathbb{P}}{\rightarrow} \sigma^2_{k, p_{\theta^0}, f^0_k, \pi^0} = \left(\sigma_k^0\right)^2$ for all $\pi^0 \in \Pi(G^0)$ and $k=1, \ldots, p$, that is 
        $$
        \frac{1}{N} \sum_{\ell = 1}^N\left(x_{\ell k} - 
        \widehat{f}_{k,  \varpi_{\pi^0}(k)}(\mathbf{x}_{\ell \varpi_{\pi^0}(k)})
        \right)^2 \overset{\mathbb{P}}{\longrightarrow} \left(\sigma_k^0\right)^2 = \mathbb{E}\left[\left(X_k - \sum_{j \in \mathbf{pa}_{G^0}(k)} f^0_{kj}(X_j)\right)^2\right].
        $$
    \end{enumerate}
    Then it holds for the derived score function $\widehat{S}$ that
    $$
    \widehat{S}(\pi^0) < \widehat{S}(\pi)
    $$
    for any $\pi^0 \in \Pi(G^0)$ and $\pi \notin \Pi(G^0)$ with probability going to $1$ for $N \rightarrow \infty$.
\end{proposition}
\paragraph{Sketch of the proof of Proposition~\ref{prop:consistency_empirical_score}}
 Our goal is to show that for any $\pi \notin \Pi(G^0)$ and $\pi^0 \in \Pi(G^0)$ it holds asymptotically 
$$\sum_{k = 1}^p \log(\widehat{\sigma}^2_{k, \widehat{f}_{k, \pi^0}}) = \widehat{S}(\pi^0) < \widehat{S}(\pi) =\sum_{k = 1}^p\log(\widehat{\sigma}^2_{k, \widehat{f}_{k, \pi}}).$$ 
By inequality~\eqref{eq:population_score_chooses_correctly} this is fulfilled if for $N \rightarrow \infty$ and $\pi \notin \Pi(G^0)$  
\begin{equation}\label{eq:score_nonoverfitting}
\lim_{N \rightarrow \infty}\widehat{S}(\pi) \geq S(\pi)    
\end{equation}
and if at the same time for $\pi^0 \in \Pi(G^0)$ it holds that
\begin{equation}\label{eq:score_consistency_correct_ordering}
\lim_{N \rightarrow \infty} \widehat{S}(\pi^0) = S(\pi^0).    
\end{equation}
Applying the continuous mapping theorem, Relation~\eqref{eq:score_consistency_correct_ordering} is ensured by Condition~2.

Contrariwise for relation~\eqref{eq:score_nonoverfitting} when $\pi \notin \Pi(G^0)$, the non-overfitting Condition~1. of Proposition~\ref{prop:consistency_empirical_score} ensures that for any $k=1, \ldots, p$ and for $N \rightarrow \infty$ 
\begin{align*}
\widehat{\sigma}^2_{k, \widehat{f}_{k, \pi}} = \frac{1}{N} \sum_{\ell =1}^N \left(x_{\ell k} - \sum_{j \in \varpi_\pi(k)}\widehat{f}_{kj, \pi}\left(x_{\ell j}\right)\right)^2
\geq \mathbb{E}_{p_{\theta^0}}\left[\left(X_{k} - \sum_{j \in \varpi_\pi(k)}\widehat{f}_{kj, \pi}\left(X_j\right)\right)^2\right].
\end{align*}
It thus follows that for $N \rightarrow \infty$
\begin{align*}
\widehat{S}(\pi) &= \sum_{k=1}^p\log\left(\widehat{\sigma}^2_{k, \widehat{f}_{k, \pi}}\right) = 
\sum_{k=1}^p\log\left(\frac{1}{N} \sum_{\ell =1}^N \left(x_{\ell k} - \sum_{j \in \varpi_\pi(k)}\widehat{f}_{kj, \pi}\left(x_{\ell j}\right)\right)^2\right) \\
& \geq \sum_{k=1}^p \log\left(\mathbb{E}_{p_{\theta^0}}\left[\left(X_{k} - \sum_{j \in \varpi_\pi(k)}\widehat{f}_{kj, \pi}\left(X_j\right)\right)^2\right]\right)
\\
&\geq \min_{(f_1, \ldots, f_p) \in \vartheta(\pi)} \sum_{k=1}^p\log(\sigma^2_{k, {p_{\theta^0}}, f_k, \pi}) = S(\pi). 
\end{align*}
A detailed proof can be found in the Appendix~\ref{sec:appendix_technical}.

%% file: 03_background.tex
As  our main result in Theorem \ref{theorem:consistency_dag_boost} relies on boosting Kernel Hilbert space regressions, we briefly introduce necessary concepts and results on boosting (Section~\ref{sec:Boosting}) and Reproducing Kernel Hilbert Spaces (RKHSs) (Section~\ref{sec:RKHS}) next. Further details on the two concepts can be found in  \cite{buehlmannYu, SchFre2012}, as well as \cite{wahba_splines,scholkopf2002learning, wainwright_hds}, respectively. 
\subsection{Boosting}\label{sec:Boosting}
$L^2$-boosting addresses the problem of finding a function $f$ in some function space $H$ that minimizes the expected $L^2$-loss
\begin{equation}\label{eq:l2_risk_population}
\frac{1}{2}\mathbb{E}_{\mathbf{X}, Y}\left[\left(Y - f(\mathbf{X})\right)^2\right].
\end{equation}
In practice, \eqref{eq:l2_risk_population} is replaced by the empirical minimizer of
\begin{equation}\label{eq:l2_risk_empirical}
\frac{1}{2N}\sum_{\ell = 1}^N \left(y_\ell - f(\mathbf{x}_\ell)\right)^2
\end{equation}
based on $N$ i.i.d.~samples $\left(\mathbf{x}_1, y_1), \ldots, (\mathbf{x}_N, y_N\right)$.
$L^2$-boosting employs a functional gradient descent approach using a base learner $S$ that maps $y^N := (y_1, \ldots, y_N)^\top$ to the estimates $\widehat{f}$ and $\widehat{y}^N = \widehat{f}(\mathbf{x}^N)$, where $\widehat{f}(\mathbf{x}^N) := \left(\widehat{f}(\mathbf{x}_1), \ldots, \widehat{f}(\mathbf{x}_N)\right)^\top$. More precisely, after initializing $\widehat{f}_0$, in each boosting step $m=1,\ldots,m_{\mbox{\scriptsize{stop}}}$ the residuals at the current $\widehat{f}^{(m)}$
$$
u_\ell = \frac{\partial}{\partial f} \frac{1}{2N} \left(y_\ell - f(x_\ell)\right)^2\vert_{f = \widehat{f}^{(m)}} = \frac{1}{N}\left(y_\ell - \widehat{f}^{(m)}(x_\ell)\right)
$$
are computed and $\widehat{f} = S(u) = S(u_1, \ldots, u_N)$ is determined.
The solution  is then used to update the estimate of the regression function 
$$\widehat{f}^{(m+1)} = \widehat{f}^{(m)} + \upsilon \widehat{f},$$
where $0 < \upsilon  \leq 1$ is the step size that is commonly fixed at a small value \citep{buehlmannYu}. 
For many base learners $S$ this leads for fixed $N$ to an overfitting if $m_{\mbox{\scriptsize{stop}}} \rightarrow \infty$. Stopping earlier is thus desired and \emph{early stopping} often applied \citep{SchFre2012}.
Following \cite{buehlmannYu, raskutti2014earlystopping}, we  consider linear and symmetric base learners $S$, that is, $S: y^N \mapsto \widehat{y}^N$ is a linear and symmetric mapping. Spline regression and linear regression, even in the generalized ridge regression sense, fall under this definition. So does the (additive) kernel ridge regression \citep{raskutti2014earlystopping, kandasamy2016additive}, which we will consider in the following. In contrast, the popular choice of decision trees are not linear. 
Proposition 1 of \citet{buehlmannYu} shows that the estimate $\widehat{\mathbf{y}}$ after $m$ boosting steps for $\mathbf{y}$ is given by 
$$
\widehat{f}^{(m)}(\mathbf{x}^N) = \widehat{\mathbf{y}}^N = B^{(m)} \mathbf{y} = (I - (I - S)^m) \mathbf{y}.
$$
As we assume $S$ to be symmetric, there exists an orthogonal $U \in \mathbb{R}^{N \times N}$ containing the eigenvectors of $S$, such that for the diagonal matrix $D$ with the eigenvalues of $S$ on the diagonal, it holds that
$
S = UDU^T.
$
It follows that
$
B^{(m)} = U(I - (I - D)^m)U^T.
$

\subsection{Reproducing Kernel Hilbert Spaces}\label{sec:RKHS}
We choose the function estimates $\widehat{f}$  from a Reproducing Kernel Hilbert Space (RKHS) $H$, while $S$ is a kernel regression estimator.
We start by introducing kernel functions.
\begin{definition}
    We call a symmetric function $K:\mathbb{R}^p \times \mathbb{R}^p \rightarrow \mathbb{R}$ a positive definite (p.d.) kernel on $\mathbb{R}^p$ if 
    $$
    \sum_{k = 1}^N \sum_{\ell = 1}^N \alpha_k \alpha_\ell K(\mathbf{x}_k, \mathbf{x_\ell}) \geq 0
    $$
    for any $\{\alpha_1, \ldots, \alpha_N\} \subset \mathbb{R}$ and any $\{\mathbf{x}_1, \ldots, \mathbf{x}_N \} \subset \mathbb{R}^p$.
\end{definition}
For any p.d.~kernel $K$, there exists a unique Hilbert space $H$ with $K(\cdot, \mathbf{x}) \in H \, \forall \mathbf{x} \in \mathbb{R}^p$ and where further it holds for any $f \in H$ and $\mathbf{x} \in \mathbb{R}^p$
\begin{equation}\label{eq:reproducing_property}
f(\mathbf{x}) = <f, K(\cdot, \mathbf{x})>_H.
\end{equation}
Equation~\eqref{eq:reproducing_property} is called the reproducing property. Consequently $H$ is called a RKHS.
By the reproducing property it holds for $f = \frac{1}{\sqrt{N}}\sum_{k = 1}^N \alpha_k K(\cdot, \mathbf{x}_k)$ and $g = \frac{1}{\sqrt{N}}\sum_{k = 1}^N \beta_k K(\cdot, \mathbf{x}_k)$, that
\begin{equation}\label{eq:inner_product_H}
<f,g>_H = \alpha^T G \beta,     
\end{equation}
where $G \in \mathbb{R}^{N \times N}$ is symmetric with $G_{jk} = \frac{K(\mathbf{x}_j, \mathbf{x}_k)}{N}$. We call $G$ the Gram matrix. By the representation theorem \citep[Proposition 12.33]{wainwright_hds} the minimizer of
\begin{equation}\label{eq:kernel_regression}
\widehat{f} = \argmin_{f \in H} \frac{1}{N} \sum_{\ell = 1}^N (y_\ell - f(\mathbf{x}_\ell))^2 + \gamma ||f||_H^2
\end{equation}
can be expressed by 
$$
\widehat{f} = \frac{1}{\sqrt{N}}\sum_{\ell=1}^N \beta_\ell K(\cdot, \mathbf{x}_\ell),
$$
where
$
\beta = \frac{1}{\sqrt{N}}(G + \gamma N I)^{-1} y^N.
$
By Equation~\eqref{eq:inner_product_H},
$
||\widehat{f}||_H^2 = \beta^T G \beta
$
holds. 
Clearly, the mapping $S: y^N \mapsto G (G + \lambda I)^{-1} y^N = \widehat{f}(\mathbf{x}^N)$ is linear and symmetric. It can be derived that $S$ has the eigenvalues 
    $
    d_\ell = \frac{\widehat{\mu_\ell}}{\widehat{\mu_\ell} + \gamma N},
    $
    where $\widehat{\mu_1}, \ldots, \widehat{\mu_N}$ are the eigenvalues of $G$. The regularization parameter $\lambda = \gamma N$ shall be constant in $N$ in this work.
    
The boosting estimate $\widehat{f}^{(m)}$ is sequentially built by adding small amounts of the current estimates. These current estimates are of the form 
$
\frac{1}{\sqrt{N}}\sum_{\ell = 1}^N \widehat{\alpha}_\ell K(\cdot, \mathbf{x}_\ell).
$
Thus, if $S$ is the base learner used for boosting, 
it holds by the construction of the boosting estimator, that there exists a $\widehat{\beta} \in \mathbb{R}^N$ with
\begin{equation}\label{eq:structure_function_estimate}
\widehat{f}^{(m)} = \frac{1}{\sqrt{N}}\sum_{\ell = 1}^N \widehat{\beta}_\ell K(\cdot, \mathbf{x}_\ell).    
\end{equation}
Here, $\widehat{f}^{(m)}$ is the boosting regression estimate after $m$ boosting steps. In this context we define
$$
\mathcal{F}_N := \left\{ f \in H: f = \frac{1}{\sqrt{N}}\sum_{\ell=1}^N \beta_\ell K(\cdot,  \mathbf{x}_\ell), ||f||_H = 1, \{\mathbf{x}_1, \ldots, \mathbf{x}_N \} \subset \mathbb{R}^p\right\}
$$
and $\widehat{f}^{(m)} \in h \mathcal{F}_N$ for some $h > 0$. 
For a continuous kernel $K \in L^2(\mathbb{R} \times \mathbb{R})$, we can define an integral operator $\mathcal{K} : L^2(\mathbb{R}) \rightarrow L^2(\mathbb{R})$ by 
$$
f(\cdot) \mapsto \left(\mathcal{K}(f)\right)(y) =  \int_\mathbb{R} K(x,y) f(x) d\mathbb{P}(x).
$$
Under assumptions on $\mathbb{P}$ and $K$, \cite{mercer_noncompact} shows that the operator $\mathcal{K}$ has eigenvalues $\mu_k \geq 0$ and eigenvectors $\phi_k \in L^2(\mathbb{R})$, so that $\mathcal{K} (\phi_k) = \mu_k \phi_k$. If the eigenvalues are ordered non-increasingly, then $\mu_k$ goes to $0$. 
The decay rate of the eigenvalues will be important in our analysis. We close this subsection with two examples.
\begin{example}[Kernel functions]\
\begin{enumerate}
    \item \textbf{Gaussian kernels on $\mathbb{R}$}: $K: \mathbb{R} \times \mathbb{R}$ is defined for some $\varsigma > 0$ by 
    $$
    K(x, x') = \exp\left(\frac{|x-x'|^2}{2\varsigma}\right).
    $$
    Its eigenvalues exist under mild assumptions on $P(\mathbf{X})$ \citep[Section 4]{mercer_noncompact} and follow an exponential decay of the form
    $$
    \mu_k \leq \exp(-C k)
    $$
    for some $C > 0$. For further details, see Section C of \cite{bach2002kernelpca} and Example 3 of \cite{cucker2002mathematical}.
    \item \textbf{Additive Kernel}: Let $H_1, \ldots, H_p$ be RKHSs with kernels $K_k$ on $X_k$, $k=1, \ldots, p$. The space $H_1 \oplus \dots \oplus H_p := \{f:\mathbb{R}^p \rightarrow \mathbb{R}: f(x_1, \ldots, x_p) = \sum_{k=1}^p f_k(x_k), f_k \in H_k\}$ is a RKHS with kernel $K = \sum_{k=1}^p K_k$. Its norm is  defined by $||f||^2_H = \sum_{k=1}^p ||f_k||^2_{H_k}$ \citep[see][Proposition 12.27]{wainwright_hds}. For Gaussian kernels $K_1, \ldots, K_p$, we assume that the eigenvalues of $K$ can be upper bounded by 
    \begin{equation}\label{eq:additive_kernel_eigenvalues}
    \mu_k \leq p\exp(-Ck)    
    \end{equation}
    for some $C > 0$. Note that for $p$ fixed, this is of type 
    $$
    \mu_k \leq \exp(-C'k)
    $$
    for some $C' > 0$. The solution of ~\eqref{eq:kernel_regression} can then be written by $\widehat{f}(x_1, \ldots, x_p) = \sum_{k=1}^p \widehat{f}_k(x_k)$, where 
    $$
    \widehat{f}_k = \sum_{\ell = 1}^N \widehat{\beta}_\ell K_k(\cdot, x_{\ell k})
    $$
    and $\widehat{\beta}$ is shared among the different components, that is, does not depend on $k$ \citep[see][for further details]{kandasamy2016additive}.
\end{enumerate}
\end{example}
The idea behind inequality~\eqref{eq:additive_kernel_eigenvalues} is the following. Each $K_k$ is a self-adjoint and compact operator for any $k=1, \ldots, p$. Let $A, B$ be linear and self-adjoint operators with  non-increasing eigenvalues $\lambda_1, \lambda_2, \ldots$ and $\mu_1, \mu_2, \ldots$, respectively. It holds by \cite{Zwahlen1965/66} for the non-increasing eigenvalues $\gamma_1, \gamma_2, \ldots$ of the self-adjoint and linear operator $A+B$ that for any $1\leq r,s\leq p$
$$
\gamma_{r + s} \leq \gamma_{r + s - 1} \leq \lambda_r + \mu_s.
$$
Now consider the non-increasing eigenvalues $\mu^k_\ell$ of the operator $A_1+ \ldots + A_k$. Let $\lambda_\ell^j, \ell = 1, 2, \ldots$ be the eigenvalues of $A_j$. Then it holds that
$$
\mu^p_{\ell p} \leq \mu^{p-1}_{(p-1)\ell} + \lambda^p_\ell  \leq \ldots \leq \lambda_\ell^1 + \ldots + \lambda_\ell^p .
$$
Inequality~\eqref{eq:additive_kernel_eigenvalues} follows under the assumption that $\lambda_\ell^j \leq \exp(-C\ell)$ for $j = 1, \ldots, p$ for some $C > 0$. 

The Gaussian (and its additive counterpart) kernel is bounded, which allows to uniformly upper-bound the supremums norm of the unit ball on $H$.
\begin{remark}\label{remark:bounded_kernel}
    If $H$ is a RKHS with kernel $K$ such that $K(\mathbf{x}, \mathbf{x}) \leq B$ for some $ B > 0$, then it holds
    $$\sup_{||f||_H \leq 1}||f||_\infty \leq B < \infty.$$
\end{remark}

%% file: 04_BoostingDAGs/04_00_boosting_dags_intro.tex
In this section we prove Theorem~\ref{theorem:consistency_dag_boost} which is our main result. It states that if we choose the regression procedure in Section~\ref{sec:CausalDiscovery} as $L^2$-boosting with early stopping, then the estimator for the topological ordering is consistent. This holds for an uniform and asymptotic number of boosting iterations. We provide the assumptions in Section~\ref{sec:Assumptions} and prove the statement in Section~\ref{sec:main_theorem}. In Section~\ref{sec:BoostingDAGs_high_dimensions} we propose an adaption of the procedure which is effective in high dimensions.

%% file: 04_BoostingDAGs/04_01_assumptions.tex
Proposition~\ref{prop:consistency_empirical_score} has shown that in order to consistently estimate the causal ordering, we have to avoid overfitting whenever we regress $X_k$ onto $\mathbf{X}_{S}$ for any $k = 1, \ldots, p$ and $S \subset \{1, \ldots, p\} \setminus \{k\}$. This poses the main challenge in applying Proposition~\ref{prop:consistency_empirical_score}. We assume that $X_k - \mathbb{E}\left[X_k | \mathbf{X}_\mathbf{S} = \mathbf{x}_\mathbf{S}\right]$ is sub-Gaussian, which later allows to control the regression estimates.
\begin{definition}
        We call a random variable $\varepsilon$ sub-Gaussian if its Orlicz norm defined by
        $$s(\varepsilon) = \inf\left\{r \in (0, \infty): \mathbb{E}\left[\exp\left(\frac{\varepsilon^2}{r^2}\right)\right] \leq 2\right\} $$ 
        is finite. 
\end{definition}
\begin{assumption}\label{ass:regression_relation_mt}
For any $k$ and $S \subset \{1, \ldots, p\} \setminus \{k\}$ consider the decomposition
$$X_k = \mu_{k, S}(\mathbf{X}_S) + \varepsilon_{k, S},$$ 
where 
$$
\mu_{k, S}(\mathbf{X}_S) = \mathbf{E}\left[X_k | \mathbf{X}_S\right] \mbox{ and } \varepsilon_{k, S} = X_k - \mathbf{E}\left[X_k | \mathbf{X}_S\right].
$$
We assume that
\begin{enumerate}
    \item $\varepsilon_{k, S} | \mathbf{X}_S = \mathbf{x}_S$ is sub-Gaussian with Orlicz norm $s_{k, S}(\mathbf{x}_S)$, $0 < s_{k, S}(\mathbf{x}_S) \leq s_{max}$ for all $\mathbf{x}_S \in \mathbb{R}^{|S|}$, and
    \item $||\mu_{k, S}||_\infty \leq \mu_{max} < \infty$.
\end{enumerate}
The constants shall hold uniformly for any $k \in \{1, \ldots, p\}$ and $S \subset \{1, \ldots, p\} \setminus \{k\}$.
\end{assumption}
We will prove that the regression estimate will lie in the function class $h \mathcal{F}_N$ for some radius $h >0$. We denote the ball of radius $h$ in $H$ by $B_h$.
In Theorem~\ref{theorem:consistency_dag_boost} we use the radius $h > 0$ and the function complexity measures Rademacher complexity and covering numbers to derive Condition~1. of Proposition~\ref{prop:consistency_empirical_score}. Both measures quantify the richness of a function class. While the Rademacher complexity of $\mathcal{F}_N$ can be upper-bounded by Theorem~\ref{theorem:rademacher_kernel} given in the Appendix, we also need to upper bound the covering numbers of $\mathcal{F}_N$. 
The complexity measures for $h \mathcal{F}_N$ can then be upper-bounded using scaling arguments. We thus make the following assumption:
\begin{assumption}\label{ass:covering_number_hilbert_space_mt}
    For any $k = 1, \ldots, p$, let $H_k$ be a RKHS on $X_k$ and $B_1^k := \{f \in H: ||f||_{H_k} \leq 1 \}$. Then it shall hold for any $z > 0$ and $k=1, \ldots, p$
    $$
    \int_0 ^1 \sqrt{\log\left( \mathcal{N}\left(\frac{uz}{2}, B^k_1\right)\right)}du < \infty.
    $$
    Here, $\mathcal{N}(\cdot, B_1^k)$ is the covering number with respect to $||\cdot||_\infty$, which is defined in Section~\ref{sec:appendix_covering}. 
    \end{assumption}
\begin{remark}
    Let $H_S = H_{s_1} \oplus \ldots \oplus H_{s_k}$ and denote the unit ball of $H_{s_k}$ by $B_1^{s_k}$ and the unit ball of $H_S$ by $B^S_1$. Assume that for any $j=1, \ldots, k$ it holds that
    $$\int_0^1 \sqrt{\log\left( \mathcal{N}\left(\frac{uz}{2}, B_1^{j}\right)\right)} du \leq C(z) < \infty$$ 
    for some $0 < C(z) < \infty$. Further, one can derive that
    $$\mathcal{N}\left(u, B^S_1\right) \leq \prod_{j=1}^k \mathcal{N}\left(\frac{u}{k}, B_1^{{s_j}}\right)
    .$$ 
    Thus, using Jensen's inequality and that $\mathcal{N}(\cdot, B_1^j)$ is non-increasing, it holds that 
    $$\int_0^1 \sqrt{\log\left( \mathcal{N}\left(\frac{uz}{2}, B^S_1\right)\right)} du \leq \int_0^1 \sqrt{\sum_{j=1}^k  \log\left( \mathcal{N}\left(\frac{uz}{2k}, B_1^{s_j}\right)\right)} du \leq \sqrt{p} C \left(\frac{z}{2p}\right) < \infty.$$
    \end{remark}
We assume that the eigenvalues of the random Gram matrix $G$ vanish with the same rate as the eigenvalues of the operator $\mathcal{K}$. For details on the connection between the eigenvalues of $G$ and $\mathcal{K}$, consult Section C of \citet{bach2002kernelpca}. 
\begin{assumption}\label{ass:decay_eigenvalues_mt}
    For $S = \{s_1, \ldots, s_k\} \subset \{1, \ldots, p\}$ let $H = H_{s_1} \oplus \ldots \oplus H_{s_k}$ be the RKHS on $\mathbf{X}_S$ generated by the additive kernel $K_{s_1} + \ldots + K_{s_k}$. There shall exist events $\mathcal{B}_N$ with $\lim_{N \rightarrow \infty} \mathbb{P}(\mathcal{B}_N) = 1$ so that on $\mathcal{B}_N$ and for $K_0 = \lfloor \frac{1}{2C_d + 1}\ln(N) \rfloor$ it holds for the empirical eigenvalues of the Gram matrix $G$ 
    $$
    \sum_{k=K_0}^N \widehat{\mu_k} \leq 
    \sum_{k=K_0}^N  \exp(-C_u k),
    $$
    and additionally 
    $$
    \widehat{\mu}_{K_0}
    \geq \exp(-C_d K_0)
    $$
    for some $C_d > C_u > 0$. The constants hold uniformly for any $S \subset \{1, \ldots, p\}$.
\end{assumption}
To give some intuition, recall that for the kernel regression estimator $S: y^N \mapsto \widehat{y}^N$ it holds that $S = U D U^\top$, where $U$ contains the eigenvectors of $G$ and $D$ is a diagonal matrix with entries $d_\ell = \frac{\widehat{\mu}_\ell}{\lambda + \widehat{\mu}_\ell}$. Clearly, $d_\ell$ has a similar decay rate as $\widehat{\mu}_\ell$. For simplicity, consider $w^N = U^\top y^N$. Assumption~\ref{ass:decay_eigenvalues_mt} then ensures that only few entries of $w^N$ largely influence $\widehat{y}^N = UD w^N$, while most entries of $w^N$ contribute little. Thus, $\widehat{y}^N$ is mostly influenced by a small subspace of $\mathbb{R}^N$.  

%% file: 04_BoostingDAGs/04_02_00boosting_dags.tex
We state now the main theorem.
\begin{theorem}\label{theorem:consistency_dag_boost}
    Let $H_{k'}$ be a RKHS on $X_{k'}$ with Gaussian kernel $K_{k'}$ for any $k' = 1, \ldots, p$. Assume that $\mathbf{X} = (X_1, \ldots, X_p)$ follows a CAM as in Assumption \ref{ass:sem_type} for functions $f^0_1 \in H_1, \ldots, f^0_p \in H_p$, where $H_k = H_{k_1} \oplus \ldots \oplus H_{k_q}$, $\left\{{k_1}, \ldots, {k_q}\right\} = \mathbf{pa}(k)$. Given Assumptions
~\ref{ass:regression_relation_mt},~\ref{ass:covering_number_hilbert_space_mt},~and~
\ref{ass:decay_eigenvalues_mt} and assuming we estimate $\widehat{f}^{({m_{stop}})}_{k, \pi} = \widehat{f}_{k, \pi} = \sum_{j \in \varpi_\pi(k)}\widehat{f}_{kj, \pi}$ using $L^2$-boosting with $X_k$ as response,  $\mathbf{X}_{\varpi_\pi(k)}$ as predictors and with number of boosting steps chosen as $m_{stop} =N^{\frac{1}{4}\frac{C_u + C_d + 1/2}{C_d + 1}}$, then it holds that 
$$\widehat{S}(\pi^0) < \widehat{S}(\pi) $$
for $N \rightarrow \infty$ with probability going to $1$ for any $\pi^0 \in \Pi^0$ and $\pi \notin \Pi^0$.
\end{theorem}
\begin{remark}
    Theorem 2 of \citet{minh_gaussian_kernel} ensures that $f_1^0, \ldots, f_p^0$ are smooth, non-constant, and non-linear and thus meet Assumption~\ref{ass:sem_type}.
\end{remark}
\begin{proof}
We apply Proposition~\ref{prop:consistency_empirical_score} and show the following two conditions.
\begin{enumerate}
    \item \emph{Boosting under Misspecification:} $\widehat{f}^{(m_{stop})}_{k, S}$ is not overfitting with respect to $(\mathbf{X}_S, X_k)$ for any $k \in \{1, \ldots, p\}$ and $S \subset \{1, \ldots, p\} \setminus \{k\}$, and
    \item \emph{Consistency of Variance Estimation:} For any $k=1, \ldots, p$ and $\pi^0 \in \Pi^0$ it holds
    $$
    \left|\frac{1}{N}\sum_{\ell=1}^N\left(\mathbf{x}_{\ell k} - \widehat{f}^{(m_{stop})}(\mathbf{x}_{\ell {\varpi_{\pi^0}(k)}}\right)^2 -\mathbb{E}\left[\left(X_k - f^0_k(\mathbf{X}_{\mathbf{pa}_{G^0}(k)}\right)^2\right] \right| \rightarrow 0
    $$
    in probability.
\end{enumerate}
We will see that 1.~follows from Theorem~\ref{theorem:boosting_misspecification} in Section~\ref{sec:boosting_misspec} and 2.~is shown in Theorem~\ref{theorem:consistency_variance} in Section~\ref{sec:variance_est_consistent}.
\end{proof}

%% file: 04_BoostingDAGs/04_02_01_boosting_under_misspecification.tex
We now show Condition $1.$ of Theorem~\ref{theorem:consistency_dag_boost}. We fix $S = \{s_1, \ldots, s_d\}$ and $k$ and define $\left(\widetilde{X}_1, \ldots, \widetilde{X}_d\right) = \widetilde{\mathbf{X}} = \mathbf{X}_S$ and $Y = X_k$.
Let $\mathbf{\widetilde{X}}^N$ be the random element containing $N$ i.i.d. observations of $\mathbf{\widetilde{X}}$ and denote the realizations of $\mathbf{\widetilde{X}}^N$ by $\mathbf{\widetilde{x}}^N = \left(\mathbf{\widetilde{x}}_1, \ldots, \mathbf{\widetilde{x}}_N \right)$. Analogously we define $Y^N$ and $y^N$.
Our goal is to prove, that
\begin{equation}\label{eq:convergence_empirical_risk}  
\left|\frac{1}{N} \sum_{\ell=1}^N \left(\widehat{f}^{(m_{stop})}(\mathbf{\widetilde{x}}_\ell) - y_\ell\right)^2 - \mathbb{E}_{\mathbf{\widetilde{X}}, Y} \left[\left(\widehat{f}^{(m_{stop})}(\mathbf{\widetilde{X}}) - Y\right)^2\right]\right| \overset{\mathbb{P}}{\longrightarrow} 0
\end{equation}
for $N \to \infty$ for the boosting estimate $\widehat{f}^{(m_{stop})}$. The left term is an expectation with respect to the empirical distribution $P_N$ (and thus depends on the realizations), whereas the right term is under the population distribution induced by $\left(\mathbf{\widetilde{X}}, Y\right)$  denoted by $P$. Thus, the l.h.s. of Equation~\eqref{eq:convergence_empirical_risk} becomes
\begin{equation}\label{eq:convergence_empirical_risk_emp_process}  
(P_N - P)\left[\left(\widehat{f}^{(m_{stop})}(\mathbf{\widetilde{X}}) - Y\right)^2\right].
\end{equation}
Assumptions \ref{ass:regression_relation},~\ref{ass:covering_number_hilbert_space},~and~\ref{ass:decay_eigenvalues} are  reformulated versions of Assumptions~\ref{ass:regression_relation_mt},~\ref{ass:covering_number_hilbert_space_mt},~and~\ref{ass:decay_eigenvalues_mt} using the notation introduced above.
\begin{assumption_prime}{\ref{ass:regression_relation_mt}}\label{ass:regression_relation}
Let 
$Y = \mu(\mathbf{\widetilde{X}}) + \varepsilon,$
for which we define the random variables
$$
\mu(\mathbf{\widetilde{X}}) = \mathbb{E}\left[Y | \mathbf{\widetilde{X}}\right] \mbox{ and } \varepsilon = Y - \mathbb{E}\left[Y | \mathbf{\widetilde{X}}\right].$$
We assume that $\varepsilon | \mathbf{\widetilde{X}} = \mathbf{\widetilde{x}}$ is sub-Gaussian with Orlicz norm $s(\mathbf{\widetilde{x}})$ and $0 < s(\mathbf{\widetilde{x}}) \leq s_{max}$ for all $\mathbf{\widetilde{x}} \in \mathbb{R}^d$ and $||\mu||_\infty \leq \mu_{max} < \infty$. Let $\sigma_{max}^2 := \max_{\mathbf{\widetilde{x}} \in \mathbb{R}^d} \mathbb{E}\left[\varepsilon^2 | \mathbf{\widetilde{X}} = \mathbf{\widetilde{x}}\right] $. 
\end{assumption_prime}

\begin{assumption_prime}{\ref{ass:covering_number_hilbert_space_mt}}\label{ass:covering_number_hilbert_space}
    Let $B_1 := \{f \in \mathcal{F}: ||f||_H \leq 1 \}$, where $H$ is the additive RKHS on $\mathbf{\widetilde{X}} = \left(\widetilde{X}_1, \ldots, \widetilde{X}_d\right)$. Then for any $z > 0$ it holds
    $$
    \int_0 ^1 \sqrt{\log\left( \mathcal{N}\left(\frac{uz}{2}, B_1\right)\right)}du < \infty.
    $$
\end{assumption_prime}
\begin{assumption_prime}{\ref{ass:decay_eigenvalues_mt}}\label{ass:decay_eigenvalues}
    There exist events $\mathcal{B}_N$ on $\mathbf{\widetilde{X}}^N$ with $\lim_{N \rightarrow \infty} \mathbb{P}(\mathcal{B}_N) = 1$ so that on $\mathcal{B}_N$ and for $K_0 = \lfloor \frac{1}{2C_d + 1}\ln(N) \rfloor $ it holds for the empirical eigenvalues of the Gram matrix $G$ for $\widetilde{\mathbf{x}}^N \in \mathcal{B}_N$
    $$
    \sum_{k=K_0}^N \widehat{\mu_k} \leq \mu_k \leq 
    \sum_{k=K_0}^N  \exp(-C_u k),
    $$
    and additionally 
    $$
    \widehat{\mu}_{K_0}
    \geq \exp(-C_d K_0)
    $$
    for some $C_d > C_u > 0$.
\end{assumption_prime}
Note that all constants in the Assumptions~\ref{ass:regression_relation},~\ref{ass:covering_number_hilbert_space},~\ref{ass:decay_eigenvalues} are independent of the choice of $k, S$. It is the purpose of this subsection to prove the following theorem (and thus Condition 1). 
\begin{theorem}\label{theorem:boosting_misspecification}
    Under the Assumptions~\ref{ass:regression_relation},~\ref{ass:covering_number_hilbert_space}~and~\ref{ass:decay_eigenvalues} it holds for $m_{stop}(N) = N^{\frac{1}{4}\frac{C_u + C_d + 1/2}{C_d + 1}}$
\begin{equation}\label{eq:convergence_p_emprirical_risk}
\left|(P-P_N) \left(Y - \widehat{f}^{(m_{stop})}(\mathbf{\widetilde{X}})\right)^2\right| \overset{\mathbb{P}}{\longrightarrow} 0.    
\end{equation}
\end{theorem}
As $\widehat{f}^{(m_{stop})}$ depends on the realizations $\mathbf{\widetilde{x}}^N$ and $y^N$, the convergence above is not trivial.
For simplicity we drop the dependency of $f, \widehat{f}^{(m_{stop})}$ on $\widetilde{\mathbf{X}}$ in the proof below.
\begin{proof}
We decompose 
\begin{align*}
&|(P_N - P)\left(Y - \widehat{f}^{(m_{stop})}\right)^2| \\
&\leq \underbrace{|(P_N - P)Y^2|}_{\rm I} + \underbrace{2|(P_N - P)\widehat{f}^{(m_{stop})}Y|}_{\rm II} + \underbrace{|(P_N - P)\left(\widehat{f}^{(m_{stop})}\right)^2|}_{\rm III}.  
\end{align*}
To prove \eqref{eq:convergence_p_emprirical_risk} we show the convergence in probability for $\rm I - III$. Term $\rm I$ goes to $0$ in probability since $Y$ has a finite fourth moment.
For term $\rm II$ it holds for any $\xi > 0$ that 
\begin{align*}
&\mathbb{P}\left(|(P_N - P) Y \widehat{f}^{(m_{stop})}| \geq  \xi\right) \\
&= \mathbb{P}\left(\left(|(P_N - P) Y \widehat{f}^{(m_{stop})}| \geq  \xi\right) \cap \left(||\widehat{f}^{(m_{stop})}||_H \in h(N)\mathcal{F}_N\right) \right)   \\
& \quad+ \mathbb{P}\left(\left(|(P_N - P) Y \widehat{f}^{(m_{stop})}| \geq  \xi\right) \cap \left(||\widehat{f}^{(m_{stop})}||_H \notin h(N)\mathcal{F}_N\right) \cap \left\{\mathbf{\widetilde{X}}^N \in \mathcal{B}_N\right\}\right)  \\
& \quad+ \mathbb{P}\left(\left(|(P_N - P) Y \widehat{f}^{(m_{stop})}| \geq  \xi\right) \cap \left(||\widehat{f}^{(m_{stop})}||_H \notin h(N)\mathcal{F}_N\right) \cap  \left\{\mathbf{\widetilde{X}}^N \notin \mathcal{B}_N\right\}\right) \\
&\leq \mathbb{P}\left(\left(|(P_N - P) Y \widehat{f}^{(m_{stop})}| \geq  \xi\right) \cap \left(||\widehat{f}^{(m_{stop})}||_H \in h(N)\mathcal{F}_N\right) \right) \\
& \quad+   \mathbb{P}\left(\left(||\widehat{f}^{(m_{stop})}||_H \notin h(N)\mathcal{F}_N\right) \cap \left\{\mathbf{\widetilde{X}}^N \in \mathcal{B}_N\right\}\right)   \\
& \quad+ \mathbb{P}\left(\left\{\mathbf{\widetilde{X}}^N \notin \mathcal{B}_N\right\}\right) \\
&\leq \mathbb{P}\left(\sup_{f \in h(N)\mathcal{F}_N}|(P_N - P) Y f| \geq  \xi\right) \\
& \quad+   \mathbb{P}\left(\left(||\widehat{f}^{(m_{stop})}||_H \notin h(N)\mathcal{F}_N\right) \cap \left\{\mathbf{\widetilde{X}}^N \in \mathcal{B}_N\right\}\right)   \\
& \quad+ \mathbb{P}\left(\left\{\mathbf{\widetilde{X}}^N \notin \mathcal{B}_N\right\}\right).
\end{align*}
with probability going to $1$ for $N \rightarrow \infty$.
For $h(N) \in o\left(N^{1/4}\right)$, the first line on the r.h.s. goes to $0$ by Corollary~\ref{corollary:covering_numbers_grow_h} and the second line vanishes by Lemma~\ref{lemma:boosting_estimate_in_scaled_spaces}. The third line converges to $0$ due to Assumption~\ref{ass:decay_eigenvalues}. This shows the convergence in probability of term $\rm II$.
\\
Similarly, for term $\rm III$ it holds for any $\xi > 0$ that
\begin{align*}
&\mathbb{P}\left(|(P_N - P) \left(\widehat{f}^{(m_{stop})}\right)^2| \geq  \xi\right) \\
&= \mathbb{P}\left(\left(|(P_N - P) \left(\widehat{f}^{(m_{stop})}\right)^2| \geq  \xi\right) \cap \left(||\widehat{f}^{(m_{stop})}||_H \in h(N)\mathcal{F}_N\right) \right)   \\
& \quad+ \mathbb{P}\left(\left(|(P_N - P)\left(\widehat{f}^{(m_{stop})}\right)^2| \geq  \xi\right) \cap \left(||\widehat{f}^{(m_{stop})}||_H \notin h(N)\mathcal{F}_N\right) \cap \{\mathbf{\widetilde{X}}^N \in \mathcal{B}_N\}\right)  \\
& \quad+ \mathbb{P}\left(\left(|(P_N - P) \left(\widehat{f}^{(m_{stop})}\right)^2| \geq  \xi\right)  \cap \left(||\widehat{f}^{(m_{stop})}||_H \notin h(N)\mathcal{F}_N\right) \cap \{\mathbf{\widetilde{X}}^N \notin \mathcal{B}_N\}\right) \\
&\leq \mathbb{P}\left(\sup_{f \in h(N)\mathcal{F}_N}|(P_N - P) f^2| \geq  \xi\right) \\
& \quad+   \mathbb{P}\left(\left(||\widehat{f}^{(m_{stop})}||_H \notin h(N)\mathcal{F}_N\right) \cap \{\mathbf{\widetilde{X}}^N \in \mathcal{B}_N\}\right)   \\
& \quad+ \mathbb{P}\left(\{\mathbf{\widetilde{X}}^N \notin \mathcal{B}_N\}\right).
\end{align*}
For $h(N) \in o\left(N^{1/4}\right)$, the first line on the r.h.s. converges to $0$ due to Corollary~\ref{corollary:uniform_convergence_norm_grow_h} 
and the other terms behave as described for term $\rm II$. This shows the convergence in probability for term $\rm III$. 
Overall, this proves Condition 1.
\end{proof}

%% file: 04_BoostingDAGs/04_02_02boosting_error_estimates_are_consistent.tex
In this paragraph we prove Condition 2. in the proof of Theorem~\ref{theorem:consistency_dag_boost}. We fix again $k$ and assume that $\varpi_{\pi^0}(k)$ has size $d$, that is, $\pi^0(k) = d+1$. We set $\mathbf{\widetilde{X}} = \left(\widetilde{X}_1, \ldots, \widetilde{X}_d\right) = \mathbf{X}_{\varpi_{\pi^0}(k)}$ and $Y = X_k$.
\begin{theorem}\label{theorem:consistency_variance}
    Let 
    $$
    Y = f^0(\mathbf{\widetilde{X}}) + \varepsilon, \quad\varepsilon \sim \mathcal{N}(0, \sigma^2),
    $$
    where $f^0$ lies in a RKHS $H$ for which Assumption~\ref{ass:decay_eigenvalues} holds with $||f^0||_H = R$.
    Then, it holds with the number of boosting steps chosen as $m_{stop} = m(N) = N^{\frac{1}{4}\frac{C_u + C_d + 1/2}{C_d + 1}}$ for $N \rightarrow \infty$
    $$
    \left|\frac{1}{N}\sum_{\ell = 1}^N (y_\ell - \widehat{f}^{(m_{stop})}(\mathbf{\widetilde{x}}_\ell))^2 - \mathbb{E}\left[\left(Y - f^0(\mathbf{\widetilde{X}})\right)^2\right] \right|= \left|\widehat{\sigma}^2 - \sigma^2 \right| \rightarrow 0 
    $$
    in probability.
\end{theorem}
\begin{proof}
We define the semi-norm 
$$
||g||^2_{2, N} := \frac{1}{N}\sum_{\ell = 1}^N g(y_\ell, \mathbf{\widetilde{x}}_\ell)^2.
$$
By the triangle inequality it holds that
\begin{equation}\label{eq:decomposition_theorem3}
    ||Y - \widehat{f}^{(m_{stop})}||_{2,N} \leq ||Y - f^0||_{2,N} + ||f^0 - \widehat{f}^{(m_{stop})}||_{2,N}.
\end{equation}
Next, we show how to asymptotically lower and upper bound $||Y - \widehat{f}^{(m_{stop})}||_{2,N}$ by $\sigma$. 

\paragraph{Lower bound:} By Remark~\ref{remark:bounded_kernel} and as $||f^0||_H = R$, $||f^0||_\infty$ is bounded. Further, $f^0(\widetilde{\mathbf{x}}) = \mathbb{E}\left[Y | \widetilde{\mathbf{X}} = \widetilde{\mathbf{x}}\right]$ and the noise $Y - f^0(\widetilde{\mathbf{x}}) = \varepsilon$ is Gaussian and thus sub-Gaussian with an Orlicz norm that is uniformly bounded.
Hence, by Theorem~\ref{theorem:boosting_misspecification} and the continuous mapping theorem the l.h.s. of~\eqref{eq:decomposition_theorem3} converges and it holds
$$||Y - \widehat{f}^{(m_{stop})}||_{2,N} = \sqrt{||Y - \widehat{f}^{(m_{stop})}||^2_{2,N}} \rightarrow \sqrt{\mathbb{E}\left[||Y - \widehat{f}^{(m_{stop})}||^2_{2}\right]} \geq \sqrt{\mathbb{E}\left[||Y - f^0||^2_{2}\right]} 
=\sigma$$ 
in probability. 

\paragraph{Upper bound:} The term $||Y - f^0||_{2,N} = \sqrt{\frac{1}{N}\sum_{\ell =1}^N \varepsilon^2_\ell}$ converges to $\sigma$ in probability. Thus, it remains to show that $||f^0 - \widehat{f}^{(m_{stop})}||_{2,N} \overset{\mathbb{P}}{\rightarrow} 0$ for $N \rightarrow \infty$, which follows from Lemma \ref{lemma:fixed_design_convergence} below.
\end{proof}

\begin{lemma}\label{lemma:fixed_design_convergence}
For $Y = f^0(\mathbf{\widetilde{X}}) + \varepsilon$, $\varepsilon \sim \mathcal{N}(0, \sigma^2)$, $f^0 \in H$, and if $\widehat{f}$ is the boosting estimate with $m_{stop} = m(N) = N^{\frac{1}{4}\frac{C_u + C_d + 1/2}{C_d + 1}}$ boosting steps, then $\widehat{f}^{(m_{stop})}$ converges to $f^0$ in a fixed design, that is,
$$
||f^0 - \widehat{f}^{(m_{stop})}||_{2,N} = \left(\frac{1}{N} \sum_{\ell = 1}^N \left(f^0(\widetilde{\mathbf{x}_\ell})- \widehat{f}^{(m_{stop})}(\widetilde{\mathbf{x}_\ell})\right)^2\right)^{1/2} \overset{\mathbb{P}}{\longrightarrow} 0.
$$
\end{lemma}
\begin{proof}
The proof is in the Appendix~\ref{sec:appendix_fixed_design_convergence}.
\end{proof}

\begin{remark}Lemma~\ref{lemma:fixed_design_convergence} also holds for heteroscedastic noise and is stated similarly in \newline \cite{buehlmannYu, raskutti2014earlystopping}.
\end{remark}
\begin{remark}
    The combination of Theorem~\ref{theorem:boosting_misspecification}~and~\ref{theorem:consistency_variance} is insightful. If we are unsure whether $f^0 \in H$ and the noise is independent,
    then limiting appropriately the number of boosting iterations leads to 
    \begin{enumerate}
        \item a consistent estimator for $f^0$ if the assumptions hold, and
        \item an estimator, so that the prediction error on the samples $\frac{1}{N}\sum_{\ell=1}^N \left(y_\ell - \widehat{f}^{(m_{stop})}(\mathbf{\widetilde{x}}_\ell)\right)^2$ is asymptotically close to the $L^2$-error $\mathbb{E}\left[\left(Y - \widehat{f}^{(m_{stop})}(\mathbf{\widetilde{X}})\right)^2\right]$ for yet unobserved realizations of $(\mathbf{\widetilde{X}}, Y)$. Here, $\frac{1}{N}\sum_{\ell=1}^N \left(y_\ell - \widehat{f}^{(m_{stop})}(\mathbf{\widetilde{x}}_\ell)\right)^2$ depends only on the observations used for learning $\widehat{f}^{(m_{stop})}$ and thus no hold-out set is necessary.
    \end{enumerate}
    We emphasize, that although the results are stated for the Gaussian kernel they can be adapted to kernels with other eigenvalue decay rates.
\end{remark}

%% file: 04_BoostingDAGs/04_03_boosting_dags_in_high_dimensions.tex
Theorem~\ref{theorem:consistency_dag_boost} shows that we can asymptotically identify the true causal ordering using boosting regressions. However, there are $p!$ possible permutations on $\{1, \ldots, p\}$ that constitute the search space. Thus, beyond a very small $p$ or without extensive prior knowledge on the topological order the computational costs are prohibitive. We address this issue through component-wise boosting in an additive noise model. 
\paragraph{Additive Noise Model}
The true graph $G^0$ and the true structural equations $f^0_1, \ldots, f^0_p$ can be represented by a function 
$F^0: \mathbb{R}^p \rightarrow \mathbb{R}^p$, where $$F^0(\mathbf{x}) = \left(f^0_1(\mathbf{x}), \ldots, f^0_p(\mathbf{x})\right)^\top = \left(f^0_1(\mathbf{x}_{\mathbf{pa}(1)}), \ldots, f^0_p(\mathbf{x}_{\mathbf{pa}(p)})\right)^\top.$$ Thus, the graph $G^0$ has an edge from $X_j$ to $X_k$ if and only the function $f_k^0$ is not constant in its $j-$th component. 
Note that the set of functions $F: \mathbb{R}^p \rightarrow \mathbb{R}^p$
corresponding to a DAG is non-convex \citep{DAGsNoTearsNonparametric}. We  assume that the structural equations $f^0_k(\mathbf{x}_{\mathbf{pa}(k)})$ decompose additively, that is~$f^0_k(\mathbf{x}_{\mathbf{pa}(k)}) = \sum_{j \in \mathbf{pa}(k)} f^0_{kj}(x_j).$

\paragraph{Component-wise Boosting}
Instead of applying $L^2$-boosting to estimate $f^0_k,\, k=1, \ldots, p$ one-by-one as in Theorem~\ref{theorem:consistency_dag_boost}, we employ component-wise boosting to estimate $F^0$.
This means, we define the loss function on the functions $F: \mathbb{R}^p \rightarrow \mathbb{R}^p$ as the log-likelihood function given $\mathbf{x}^N$
$$
L(F, \mathbf{x}^N) = L\left( \left(f_1, \ldots, f_p\right), \mathbf{x}^N\right) = \sum_{k=1}^p \log\left(\sum_{\ell = 1}^N  \left(\mathbf{x}_{\ell k} - f_k(\mathbf{x}_\ell)\right)^2 \right)
$$
and proceed as follows. Choose a step size $0 < \mu \leq 1$ and let $F^{(1)}=0$ be the starting value. Then, for $m=1,2,\ldots$ we set $f_k^{(m + 1)}(\mathbf{x}) =  f_k^{(m)}(\mathbf{x}) + \mu \widehat{f}_{kj}(x_j)$, where $(j,k) \in \{1, \ldots, p\} \times \{1, \ldots, p\}$ is the solution of
$$
\argmin_{(j,k) \notin N^m} S(j, k; F^{(m)}),
$$
and $S(j,k; F^{(m)})$ is the score on the edges defined by
\begin{equation}\label{eq:score_dagboost_large_p}
S(j,k; F^{(m)}) := \log\left(\sum_{\ell = 1}^N \left(\widehat{f}_{kj}(\mathbf{x}_{\ell j}) - \left(\mathbf{x}_{\ell k} - f_k^{(m)}(\mathbf{x_\ell}) \right) \right)^2 \right).
\end{equation}
Here, the candidate functions $\widehat{f}_{kj}$ are determined by solving the kernel ridge regression
\begin{equation}\label{eq:candidate_functions_large_p}
\widehat{f}_{kj} = \argmin_{g_{kj} \in H_j} \sum_{\ell = 1}^N \left(g_{kj}(\mathbf{x}_{\ell j}) - \left(\mathbf{x}_{\ell k} - f_k^{(m)}(\mathbf{x}_\ell) \right)\right)^2 + \lambda||g_{kj}||^2_{H_j}.    
\end{equation}
In the set $N^m$ we track the edges $(j,k)$ that would cause a cycle when added to $F^{(m)}$. Hence, $F^{(m)}$ corresponds to a DAG for any $m=1, 2, \ldots$.

Note that if the edge $(j,k)$ is chosen,
then we only need  to update $S(j,k; F^{(m+1)})$ for $(j,k) \notin N^{m+1}$, while this number remains unchanged for $k' \neq k$, that is, $S(j,k'; F^{(m+1)}) = S(j,k'; F^{(m)})$. This reduces the computational burden.
We stop the procedure after $m_{stop}$ steps, which is the crucial tuning parameter of (component-wise) boosting.
\paragraph{Choosing $m_{stop}$}
Inspired by results from boosting for regression \citep{tutz_binder_boosting_aic, hothorn_buehlmann_aic} we use the Akaike Information Criterion (AIC) to select $m_{stop}$. For any $f^{(m)}_k$ we calculate the trace of the mapping $B^{(m)}_k: \left(x_{1 k}, \ldots, x_{N k}\right) \mapsto \left(f^{(m)}_k(\mathbf{x}_{1}), \ldots, f^{(m)}_k(\mathbf{x}_{N})\right)$. Then we define the AIC score
    \begin{equation}\label{eq:AIC}
    AIC(F^{(m)}, \mathbf{x}^N) = \sum_{k=1}^p AIC_k(f_k^{(m)}, \mathbf{x}^N) = \sum_{k=1}^p \left(\sum_{\ell = 1}^N \left(x_{\ell k} - f^{(m)}_k(\mathbf{x}_{\ell})\right)^2 +tr(B^{(m)}_k) \right).
    \end{equation}
We stop the procedure and set $m_{stop} = m$ if $AIC(F^{(m)}, \mathbf{x}^N)$ increases with $m$. We emphasize that this is merely a local minimum w.r.t~to the AIC score and the global optimum is hard to find due to the non-convexity of the search space.
The algorithm using the AIC is outlined in Algorithm~\ref{algo:dagboost_large_p} in Appendix~\ref{sec:appendix_algorithm}. It can be understood as a component-wise functional gradient descent in the space of those additive functions $\mathbb{R}^p \rightarrow \mathbb{R}^p$ which imply a DAG. 
\paragraph{Pruning}
We prune the estimated graph $\widehat{G}$ by running an additive model regression of every node on its parents. Finally, we keep only those nodes as parents whose $p$-value is below $0.001$. For more details on pruning, see \cite{CAM}.

%% file: 05_SimulationStudy.tex
In this section we empirically investigate the proposed algorithms. In Section~\ref{sec:data_generation} we describe the data-generating processes. 
In Section~\ref{sec:simstudy_small_p} we verify Theorem~\ref{theorem:consistency_dag_boost} for data sets of small dimensions. 
In Section~\ref{sec:simstudy_large_p} we benchmark the algorithm of Section~\ref{sec:BoostingDAGs_high_dimensions} on high-dimensional data sets against state-of-the-art methods. Further, we run a sensitivity analysis on the effect of the hyperparameters.

Every presented result is based on $100$ randomly generated data sets unless stated otherwise. Aside from the sensitivity analysis, we set the step size $\mu = 0.3$ and the penalty parameter $\lambda = 0.01$. While it is known that boosting is commonly robust with respect to the step size (as long as it is small enough), we find in Section \ref{sec:sensitivity_analysis}, that our method is robust also against the specific choice of $\lambda$. We therefore refrain from further tuning here.

\subsection{Sampling SEMs and Data Generation}\label{sec:data_generation}
The generation of the synthetic data sets follows closely \cite{CAM, lachapelle2019gradient, DAGsNoTearsNonparametric, masked_dag_learning}, that is, they are generated as follows. 
\begin{enumerate}
    \item Generate \textbf{underlying graph} $G^0$ with one of the following two methods.
    \begin{enumerate}
        \item Generate a DAG according to the Erdös-Renyi (ER) model \citep{erdos59a}. That means, we first generate a random DAG with the maximal number of edges and keep every edge with a constant probability. Every node has the same distribution for the number of its neighbors.
        \item Generate a scale-free (SF) graph using the model of \cite{barabasi}. There exist hubs of nodes with large degree, while other nodes have a smaller degree. This graph structure is observed in various applications \citep{jeong2000large,wille2004sparse,  kertel2022learning}.
    \end{enumerate}
    \item Generate \textbf{structural equations} (SEs) in one of the two ways.
    \begin{enumerate}
    \item \textbf{Additive:} For any edge $(j,k)$ in the graph, sample $f^0_{kj}$ from a Gaussian process with mean $0$ and covariance function $cov(f^0(x_{\ell j}), f^0(x_{\ell' j})) = \exp\left(- \frac{(x_{\ell j} - x_{\ell' j})^2}{2}\right)$ and set $f^0_k(\mathbf{x}_{\ell \mathbf{pa}_{{G^0}}(k)}) = \sum_{j \in \mathbf{pa}_{G^0}(k)} f^0_{kj}(x_{\ell j})$ for $\ell, \ell' = 1, \ldots, N$.
    \item \textbf{Non-additive:} For every node $k$ consider its parents $\mathbf{pa}_{G^0}(k)$ and sample $f^0_k$ from a Gaussian Process with mean $0$ and covariance function 
    $$cov(f^0(x_{\ell \mathbf{pa}_{{G^0}}(k)}), f^0(x_{\ell' \mathbf{pa}_{{G^0}}(k)})) = \exp\left(-\frac{||x_{\ell \mathbf{pa}_{{G^0}}(k)} - x_{\ell' \mathbf{pa}_{{G^0}}(k)}||_2^2}{2}\right),$$
    for $\ell, \ell' = 1, \ldots, N$.
    \end{enumerate}
    \item Sample the \textbf{standard deviation} $\sigma^0_k$ for all $k = 1, \ldots, p$ from a uniform distribution on $\left[\sqrt{2}/5, \sqrt{2}\right]$.
\end{enumerate}
Finally, we sample the variables with no incoming edges from a centered Gaussian distribution with standard deviation as chosen in step 3. Following the topological order of $G^0$, we generate the data set recursively according to the SEM. We emphasize, that non-additive SEs conflict with Assumption~\ref{ass:sem_type}.

\subsection{Low-Dimensional Data}\label{sec:simstudy_small_p}
In this low-dimensional simulation study we generate data by setting $p=5$ and sample from an ER graph with on average five edges.

Then, we calculate the score for any of the $p!$ permutations $\pi$ as described in Section~\ref{sec:CausalDiscovery}. For every regression we choose $m_{stop}$ to be the minimal $m$, for which $AIC_k(f_k^{(m)}, \mathbf{x}^N)$
increases with $m$. Recall that $f^{(m)}_k$ is a function on $\mathbf{x}_{\varpi_\pi(k)}$ and we use regular (not component-wise) boosting.
We generate data sets with $N=10, 20, 50, 100, 200$ observations with either additive or non-additive SEs. To evaluate the quality of the estimated permutations, we use the transposition distance
$$
d_{trans}(\pi_1, \pi_2) := \min \left| \{ \text{transpositions } \sigma_1, \ldots, \sigma_J: \sigma_1 \circ \ldots \circ \sigma_J \circ \pi_1 = \pi_2 \}  \right|.
$$
For an estimated permutation $\widehat{\pi}$ we then set 
$$
d_{trans}\left(\widehat{\pi}, \Pi^0\right) := \min_{\pi^0 \in \Pi^0} d_{trans}(\widehat{\pi}, \pi^0).
$$
Thus, we calculate the minimal number of adjacent swaps so that the estimated permutation aligns with the underlying topological order. 
\begin{table}[h!]
    \centering
    \begin{tabular}{c|cc|cc}
         & \multicolumn{2}{c}{Additve SEs} & \multicolumn{2}{c}{Non-additve SEs} \\
         $N$ & $\overline{\left(d_{trans}(\widehat{\pi}, \Pi^0)\right)}$ & SD$\left(d_{trans}(\widehat{\pi}, \Pi^0)\right)$ & $\overline{\left(d_{trans}(\widehat{\pi}, \Pi^0)\right)}$ & SD$\left(d_{trans}(\widehat{\pi}, \Pi^0)\right)$ \\
         \hline
         10 & 3.29 & 1.90 & 3.26 & 1.95  \\
         20 & 1.76 & 1.93 & 2.03 & 2.00 \\
         50 & 0.49 & 1.12 & 0.85 & 1.55 \\
         100 & 0.35 & 0.92 & 0.56 & 1.04 \\
         200 & 0.03 & 0.22 & 0.23 & 0.61
    \end{tabular}
    \caption{Mean ($\overline{\left(d_{trans}(\widehat{\pi}, \Pi^0)\right)}$) and standard deviation (SD) of the transposition distances ($d_{trans}$) between the estimated permutation $\widehat{\pi}$ and the set of true permutations $\Pi^0$ for SEMs with additive and non-additive SEs. The underlying graphs are of ER type with on average five edges. }
    \label{tab:small_p}
\end{table}
The results are depicted in Table~\ref{tab:small_p} and as expected, increasing the sample size decreases the transposition distances. This supports Theorem~\ref{theorem:consistency_dag_boost}. Further, the convergence seems also to hold under non-additive SEs. This indicates a robustness of the algorithm with respect to non-additve SEs.

\subsection{High-
Dimensional Data}\label{sec:simstudy_large_p}
\subsubsection{Comparison with Existing Methods}
In the following we compare the method proposed in Section~\ref{sec:BoostingDAGs_high_dimensions} (denoted by DAGBoost) with \cite{CAM} (denoted by CAM)  and \cite{DAGsNoTearsNonparametric} (denoted by DAGSNOTEARS). For CAM we employ the default configuration  and for DAGSNOTEARS we set $\lambda = 0.03$ and cutoff to $0.3$. For a comparison with other methods see \cite{CAM}. 

As performance measure we calculate the Structural Hamming Distance (SHD) between the true and the estimated graph for data sets containing $N=200$ observations of $p=100$ variables. The mean and standard deviation (SD) of the SHDs are given in Table~\ref{tab:large_p_results}.

 \begin{table}[ht]
    \centering
    \begin{tabular}{cc|cc|cc|cc}
         &   & \multicolumn{2}{c}{CAM}  & \multicolumn{2}{c}{DAGBoost} & \multicolumn{2}{c}{DAGSNOTEARS}  \\
         Additive & Graph   & $\overline{\text{SHD}}$ & SD(SHD)  & $\overline{\text{SHD}}$ & SD(SHD) & $\overline{\text{SHD}}$ & SD(SHD) \\
         \hline
         True & SF  & 30.18 & 9.47 & 30.08 & 9.57 & 176.96 & 23.06\\
         True & ER  & 12.07  & 4.24  & 14.90 & 7.12 & 127.64 & 15.84 \\
         \hline
         False & SF  &  77.63 & 7.57 & 67.47 & 5.73 & 135.32 & 18.20\\
         False & ER & 36.40 & 7.96 & 37.57 & 9.15 & 111.78 & 13.16\\
    \end{tabular}
    \caption{Mean ($\overline{\text{SHD}}$) and standard deviation (SD) of SHDs between the true graph and the graphs estimated by the three presented algorithms.}
    \label{tab:large_p_results}
\end{table}

From this table we make the following observations.  DAGSNOTEARS does not provide satisfying results, while CAM and DAGBoost perform noticeably better in all simulation scenarios. Compared to DAGBoost, CAM achieves slightly better results for the easiest setting of ER graphs and additive SEs (improvement by $0.5$ standard deviations). Generally, all methods suffer from an uneven edge distribution (SF graphs). However, for DAGBoost the mean of the SHDs increases less than CAM when the graph is of type SF instead of ER.
Thus, in the most complex scenario of non-additve SEs and a non-even distribution of the edges among the nodes (SF graphs), DAGBoost is more than one standard deviation better than CAM. This is an important insight for real-world applications, which often follow SF graphs. 
\begin{table}[ht]
    \centering
    \begin{tabular}{cc|cc|cc|cc}
         &   & \multicolumn{2}{c}{CAM}  & \multicolumn{2}{c}{DAGBoost} & \multicolumn{2}{c}{DAGSNOTEARS}  \\
         Additive & Graph   & Precision & Recall  & Precision & Recall & Precision & Recall \\
         \hline
         True & SF  & 0.868  &0.817 & 0.967 & 0.712  & 0.165 & 0.198 \\
         True & ER  &  0.907  & 0.979 & 0.933 & 0.920 & 0.174 & 0.154 \\
         \hline
         False & SF  &  0.676 & 0.383 & 0.930 & 0.345   & 0.037  & 0.024\\
         False & ER & 0.823 & 0.788  & 0.918 & 0.693 & 0.127 & 0.083 \\
    \end{tabular}
    \caption{Mean of precision and recall for the three presented algorithms. Precision is the ratio between the correctly identified edges and all identified edges. Recall is the share of the correctly identified edges among all true edges. }
    \label{tab:large_p_precision_recall}
\end{table}

Table~\ref{tab:large_p_precision_recall} furthermore summarizes the mean of the precision and the recall for the three algorithms. The precision, that is, the ratio of the correctly identified edges to all identified edges, is larger for DAGBoost. On the other hand, the recall, which is the share of the identified edges among all true edges, is larger for CAM. Hence, the edges of DAGBoost are more reliable, while CAM misses a lower number of the underlying relationships.

Overall, DAGBoost is a strong competitor to CAM, both of which outperform \newline DAGSNOTEARS. In particular, DAGBoost tends to estimate more reliable edges and is superior compared to CAM in complex non-additive and SF settings. 
Last, we make note the advantage of less tuning required for DAGBoost compared to CAM. 
While CAM has additional tuning parameters for the preliminary neighborhood selection \citep[see][]{CAM}, the tuning parameters of DAGBoost are merely the step size $\mu$ and the penalty parameter $\lambda$ which we found to be rather robust (see the sensitivity analysis below), thus relatively easy to tune. Pruning the resulting graph has a positive effect on the SHD between the estimated and the true graph for DAGBoost. However, this effect is much larger for CAM as the graph before pruning contains many more edges. Thus, DAGBoost is less reliant on the hyperparameters of the pruning step compared to CAM.

\subsubsection{Sensitivity Analysis}\label{sec:sensitivity_analysis}
We investigate the performance of DAGBoost with respect to a variation of the step size $\mu$ and the regularization parameter $\lambda$. 

When varying $\lambda$ we set $\mu = 0.3$. On the other hand when varying $\mu$ we fix $\lambda = 0.01$.
We conduct our analysis with ER graphs with $p=100$ nodes and additive SEs. The mean and standard deviation of the SHD between the estimated and true graph are reported in Tables~\ref{tab:sensitivity_mu}~and~\ref{tab:sensitivity_lambda}.

\begin{table}[ht]
    \centering
    \begin{tabular}{c|cc|cc}
         $\mu$ & mean(SHD)  & SD(SHD) & mean(runtime in s) & SD(runtime in s) \\
         \hline
         0.3 & 14.90 &  7.12 & 58.47 & 14.82\\
         0.5& 14.21 & 7.17 & 43.72 & 10.97 \\
         0.7&  13.65  & 7.13 & 37.45 & 7.46\\
         0.9& 13.17 & 7.05 &  20.17 & 2.84\\
    \end{tabular}
    \caption{Mean and SD of SHDs between estimated and true graph for a varying step size $\mu$. The penalty parameter is fixed at $\lambda = 0.01$. The graphs are of ER type and the SEs are additive. The runtime statistics are based on $10$ experiments.}
    \label{tab:sensitivity_mu}
\end{table}
\begin{table}[ht]
    \centering
    \begin{tabular}{c|cc|cc}
         $\lambda$ & mean(SHD)  & SD(SHD) & mean(runtime in s) & SD(runtime in s) \\
         \hline
         0.001 & 15.58 & 7.56 & 43.54 & 10.23 \\
         0.01& 12.12 & 6.92 & 64.10 & 19.28 \\
         0.1& 14.90 & 7.12 & 204.87 & 63.54 \\
    \end{tabular}
    \caption{Mean and SD of SHDs between estimated and true graph for a varying penalty parameter $\lambda$. The step size is fixed at $\mu = 0.3$. The graphs are of ER type and the SEs are additive.
    The runtime statistics are based on $10$ experiments.}
    \label{tab:sensitivity_lambda}
\end{table}

One can see that the influence of the hyperparameters on the SHDs is minor.
The AIC score controls the number of boosting steps $m_{stop}$ very efficiently. 
It thus accounts well for the different base learners, which depend on the step size $\mu$ and the penalty parameter $\lambda$.
An increase in the step size or a reduction in the penalty parameter leads to a smaller number of boosting iterations which in turn leads to a reduced runtime. At the same time the quality of the estimated graph is not strongly effected. We thus recommend to use DAGBoost with a large step size or a low penalty parameter if the computational resources or the time are limited.  

Although the impact of the hyperparameters is shown for one specific setting, based on our observations, they similarly hold in a wide range of data-generating processes.

%% file: 06_conclusion.tex
In this work we investigated boosting for causal discovery and for the estimation of the causal order. We proposed a generic score function on the orderings that depends on a regression estimator. We presented two sufficient conditions on the regression estimator, so that the score function can consistently distinguish between aligning and incompatible orderings. 
\begin{enumerate}
    \item The regression estimator must consistently find the true regression function in a correctly specified scenario with homoscedastic noise, and
    \item the mean squared prediction error on the samples must converge to the expected $L^2$-prediction error for yet unseen observations even in general misspecified scenarios.
\end{enumerate}
Together, the conditions imply a safety net for the regression estimator, which is interesting on its own. In a misspecified setting, the fit of the regression function to the observed samples gives a good estimate for the expected squared prediction error for yet unobserved realizations. In a correctly specified setting on the other side, the regression estimator still identifies the underlying functional relationship.

We  showed that boosting with appropriate early stopping provides this safety net. Thus, our analysis gives insights on the generalization ability of boosting procedures for real-world data, which most likely does not meet all model assumptions. 

In order to use a score function on the orderings for the identification of the topological order, one needs to score every possible permutation. 
This is infeasible for large $p$ and insufficient prior knowledge on the causal structure. Thus we proposed a greedy boosting in the space of functions of $\mathbb{R}^p \rightarrow \mathbb{R}^p$ which correspond to a DAG. The algorithm can be understood as a functional gradient descent in the space of additive SEMs, aka component-wise boosting.

A simulation study underlined that the score function on the permutations consistently prefers a correct causal order and the convergence manifests already for small $N$. 
These findings were even robust to non-additive SEs. For small $p$ or in case of extensive prior knowledge that drastically reduces the search space of permutations, the combination of the score and a feature selection procedure can be used for deriving the causal graph. 

The second part of our simulations study showed that the gradient descent is highly competitive with state-of-the-art algorithms. Particularly for complex data-generating processes, the algorithm provides a noticeable benefit. Besides, the exact choice of the hyperparameters are efficiently and automatically balanced by the number of boosting iterations and the AIC score. Thus, the procedure is easy to tune and ready to be applied to a variety of data sets.

Many parts of our analysis were generic and the RKHS regression can easily be replaced by other regression estimators as spline regression or neural networks. Thus one can further investigate which regression estimators lead to consistent estimators for the causal order and the empirical performance and theoretical properties are to be explored. A combination of gradient-based methods as in \cite{DAGsNoTears} and boosting could also lead to insights.

%% file: Y_Appendix/A_proof_prop_1.tex
\begin{proof}
Recall that $\varpi_\pi(k) := \{j: \pi(j) < \pi(k)\}$.
    \begin{align*}
        \widehat{S}(\pi) &= \sum_{k=1}^p \log(\widehat{\sigma}^2_{k, \widehat{f}_{k, \pi}})  \\
        &\geq \min_{(f_1, \ldots, f_p) \in \vartheta(\pi)} \sum_{k=1}^p \log\left((P_N - P) \left(X_k - \widehat{f}_{k, \pi}(\mathbf{X}_{\varpi_\pi(k)})\right)^2 + \sigma^2_{k, p_{\theta^0}, f_k, \pi}\right) \\
        &\geq \min_{(f_1, \ldots, f_p) \in \vartheta(\pi)} \sum_{k=1}^p \log(\sigma^2_{k, p_{\theta^0}, f_k, \pi}) \\
        &\quad+ \underbrace{\max\left\{0, \frac{-(P_N - P) \left(X_k - \widehat{f}_{k, \pi}(\mathbf{X}_{\varpi_\pi(k)})\right)^2}{ \sigma^2_{k, {p_{\theta^0}}, f_k, \pi} + (P_N - P) \left(X_k - \widehat{f}_{k, \pi}(\mathbf{X}_{\varpi_\pi(k)})\right)^2}\right\}}_{=:\Delta_{N, k}} \\
        &= \min_{(f_1, \ldots, f_p) \in \vartheta(\pi)} \sum_{k=1}^p \log(\sigma^2_{k, p_{\theta^0}, f_k, \pi}) + \sum_{k=1}^p \Delta_{N, k} \\
        &= S(\pi) + \sum_{k=1}^p \Delta_{N, k} \\
        &= S(\pi^0) + \sum_{k=1}^p\Delta_{N, k} + \underbrace{S(\pi) - S(\pi^0)}_{\xi_{\pi, \pi^0} > 0}  \\
        &= \sum_{k=1}^p\log\left(\left(\sigma_k^0\right)^2 - \widehat{\sigma}^2_{k, \widehat{f}_{k, \pi^0}} + \widehat{\sigma}^2_{k, \widehat{f}_{k, \pi^0}} \right)+\Delta_{N,k} + \xi_{\pi, \pi^0}\\
        &\geq \sum_{k=1}^p \log\left(\widehat{\sigma}^2_{k,\widehat{f}_{k, \pi^0}}\right) +\underbrace{\max\left\{0, -\frac{\left(\sigma_k^0\right)^2 - \widehat{\sigma}^2_{k,\widehat{f}_{k, \pi^0}}}{\left(\sigma_k^0\right)^2}\right\}}_{=:\gamma_{N, k}} + \Delta_{N,k} + \xi_{\pi, \pi^0} \\
        &= \widehat{S}(\pi^0) + \sum_{k=1}^p \left(\gamma_{N, k} + \Delta_{N,k} \right) + \xi_{\pi, \pi^0}
    \end{align*}
    In the first inequality, we used that for any $k=1, \ldots, p$
    \begin{align*}
    P\left((X_k - \widehat{f}_{k, \pi}(\mathbf{X}_{\varpi_\pi(k)})^2\right) 
    &= \mathbb{E}_{p_{\theta^0}}\left[(X_k - \widehat{f}_{k, \pi}(\mathbf{X}_{\varpi_\pi(k)})^2\right] \\
    &\geq   \min_{(f_1, \ldots, f_p) \in \vartheta(\pi)} \mathbb{E}_{p_{\theta^0}}\left[(X_k - f_{k}(\mathbf{X}_{\varpi_\pi(k)})^2\right] \\
    &=  \min_{(f_1, \ldots, f_p) \in \vartheta(\pi)}\sigma^2_{k, p_{\theta^0}, f_k, \pi}.
    \end{align*}
In the second and last inequality Lemma~\ref{lemma:log_inequality} below was used. Further, $\xi_{\pi, \pi^0} > 0$, which does not depend on $N$,  by the identifiability of the model. By the assumptions and the continuous mapping theorem it holds
    \begin{equation*}
   \mathbb{P}\left( |\Delta_{N, k}| \geq \frac{\xi_{\pi, \pi^0}}{2p}\right) \to 0 \mbox{ and }\mathbb{P}\left( |\gamma_{N, k}| \geq \frac{\xi_{\pi, \pi^0}}{2p}\right) \to 0 \mbox{ for all } k = 1, \ldots, p \mbox{ and } N\to\infty,
    \end{equation*}
    from which we derive that 
    $$
   \xi_N= |\sum_{k = 1}^p \Delta_{N, k} + \gamma_{N, k}| \leq \sum_{k = 1}^p |\Delta_{N, k}| + \sum_{k = 1}^p|\gamma_{N, k}| < \xi_{\pi, \pi^0}
    $$
    with probability going to $1$ for $N \rightarrow \infty$.
\end{proof}

\begin{lemma}\label{lemma:log_inequality}
    For $x > 0$ and $x + \delta > 0$ it holds that
    $$
    \log(x + \delta) \geq \log(x) + \max\left\{0, -\frac{\delta}{x + \delta}\right\}
    $$
\end{lemma}
\begin{proof}
    The statement is true for $\delta \geq 0$. For $\delta < 0$ it holds that
    \begin{align*}
    \log(x) &= \log(x - |\delta| + |\delta|) \leq \log(x - |\delta|) + \frac{|\delta|}{x - |\delta|} = \log(x + \delta) - \frac{\delta}{x + \delta} \\ 
    &=  \log(x + \delta) - \max\left\{0, -\frac{\delta}{x + \delta}\right\}    
    \end{align*}
    from which the result follows.
\end{proof}

%% file: Y_Appendix/B_proof_theorem2.tex
The proof of Theorem~\ref{theorem:boosting_misspecification} decomposes $(P_N-P) \left(Y-\widehat{f}^{(m_{stop})}\right)^2$ into $(P_N - P) \left(Y\widehat{f}^{(m_{stop})}\right)$ and $(P_N-P) \left(\widehat{f}^{(m_{stop})}\right)^2$. We show both convergences using the theory of empirical processes. 

For this purpose, we show in Section~\ref{sec:appendix_upperbound_rkhs_norm} that $||\widehat{f}^{(m_{stop})}||_H$ can be upper-bounded. Section~\ref{sec:appendix_covering} then derives the convergence of $(P_N - P) \left(Y\widehat{f}^{(m_{stop})}\right)$ using \textit{Covering Numbers}. The convergence of $(P_N - P) \left(\widehat{f}^{(m_{stop})}\right)^2$ is then shown using \textit{Rademacher Complexities}. 

%% file: Y_Appendix/B1_Upperbound_h_norm.tex
The main result of this section is Lemma~\ref{lemma:boosting_estimate_in_scaled_spaces}, which upper-bounds the Hilbert space norm of the boosting estimate $||\widehat{f}^{(m)}||_H$ with a probability going to $1$ for $N \rightarrow \infty$ for a suitably chosen number of boosting iterations $m$. The analysis is based on the fact that $||\widehat{f}^{(m)}||_H^2$ can be expressed as a quadratic form.
\begin{lemma}\label{lemma:expression_h_norm}
    Let $S$ be the kernel regression learner with penalty parameter $\lambda$. Assume that the Gram matrix $G$ is invertible. Then it holds for the boosting estimate after $m$ boosting steps that
    $$
    ||\widehat{f}^{(m)}||_H^2 = \frac{1}{N} ({y^N})^T U (I - (I - D)^m)^2 \Lambda^{-1} U^T {y^N},
    $$
    where $U, D, \Lambda \in \mathbb{R}^{N \times N}$ and $D, \Lambda $ are diagonal matrices. $\Lambda$ has the eigenvalues $\widehat{\mu}_1, \ldots, \widehat{\mu}_N$ of $G$ and $D$ has $\frac{\widehat{\mu}_1}{\widehat{\mu}_1 + \lambda}, \ldots, \frac{\widehat{\mu}_N}{\widehat{\mu}_N + \lambda}$ on the diagonal. $U$ contains the corresponding eigenvectors of $G$.
\end{lemma}
\begin{proof}
It holds for the linear base learner $S$ mapping $y^N$ to $\widehat{f}(\mathbf{x}^N)$, that
$$
S = G(G + \lambda I )^{-1}.
$$
The matrix $S$ is symmetric and  has the eigenvalues $d_1 = \frac{\widehat{\mu}_1}{\widehat{\mu}_1 + \lambda}, \ldots, d_N = \frac{\widehat{\mu}_N}{\widehat{\mu}_N + \lambda}$. Thus, for the orthogonal matrix $U$ containing the eigenvectors of $S$ and the diagonal matrix $D$ containing the eigenvalues $d_1, \ldots, d_N$ of $S$ it holds that 
$$S = U D U^T.$$
By Equation~\eqref{eq:structure_function_estimate}, the estimate $\widehat{f}^{(m)} = B^{(m)} y^N = (I - (I - S)^m)y^N$ can be expressed by 
$$
\widehat{f}^{(m)}(\mathbf{\widetilde{x}}) = \frac{1}{\sqrt{N}}\sum_{k=1}^N \widehat{\beta}_k K(\mathbf{\widetilde{x}}, \mathbf{\widetilde{x}}_k)
$$
for some $\widehat{\beta} \in \mathbb{R}^N$. Using the results of Section~\ref{sec:RKHS} we obtain 
\begin{align*}
    ||\widehat{f}^{(m)}||_H^2 
    &= \widehat{\beta}^\top G \widehat{\beta} \\
    &= \widehat{f}^{(m)}(\widetilde{x}^N)^\top \frac{G^{-1}}{\sqrt{N}} G \frac{G^{-1}}{\sqrt{N}}\widehat{f}^{(m)}(\widetilde{x}^N) \\
    &= \frac{1}{N} ({y^N})^\top B^{(m)} G^{-1} B^{(m)} {y^N}.
\end{align*}
Note that $G$, $S$ and $B^{(m)}$ have the same eigenvectors. Besides, $\Lambda$ and $I - (I-D)^m$ are diagonal matrices, so they commute. Hence, $B^{(m)} G^{-1} B^{(m)} = U (I - (I-D)^m) U^\top U \Lambda^{-1} U^\top U (I - (I-D)^m) U^\top =  U (I - (I - D)^m)^2 \Lambda^{-1}U^T$. Thus, 
$$
||\widehat{f}^{(m)}||_H^2 = \frac{1}{N} ({y^N})^\top B^{(m)} G^{-1} B^{(m)} {y^N} = \frac{1}{N} ({y^N})^\top U (I - (I - D)^m)^2 \Lambda^{-1} U^\top {y^N}.
$$
\end{proof}
The following Hanson-Wright-inequality gives probabilistic upper bounds for quadratic forms as derived in Lemma~\ref{lemma:expression_h_norm}.
\begin{theorem}[Hanson-Wright-Inequality]\cite[Theorem 1.1]{hansonwright_inequality}\label{theorem:hanson_wright}
Consider a random vector $\mathbf{Z} = (Z_1, \ldots, Z_N) \in \mathbb{R}^N$ with independent components, for which $\mathbb{E}(Z_\ell) = 0, \ell = 1, \ldots, N$ and for which the Orlicz norm of $Z_1, \ldots, Z_N$ is uniformly bounded by $s_{max}$. For any $M \in \mathbb{R}^{N \times N}$ it holds for every $t>0$
$$
\mathbb{P}\left( \left||| M \mathbf{Z}||^2 - \mathbb{E}\left[|| M \mathbf{Z}||^2\right] \right|> t\right) \leq 2 \exp\left(-c \min\left\{ \frac{t^2}{s_{max}^4 ||M^2||_F^2}, \frac{t}{s_{max}^2 ||M^2||}\right\}\right).
$$
\end{theorem}
Lemma~\ref{lemma:expression_h_norm} shows that the RKHS norm can be expressed as a quadratic form. In the next steps, we upper bound the quadratic form, that is, the RKHS norm of $\widehat{f}^{(m)}$ using Theorem~\ref{theorem:hanson_wright}. We see, that we can choose $M$ in Theorem~\ref{theorem:hanson_wright} as the matrix square root of $U (I - (I - D)^m)^2 \Lambda^{-1} U^T$ so that $||MY||^2 = ||\widehat{f}^{(m)}||_H^2$. Thus, these bounds depend on the number of boosting steps $m$ and $D$ and $\Lambda$ (which are functions of $G$), where the latter are probabilistically depending on $\mathbf{\widetilde{X}}^N$. Controlling $D$, this allows us to vary $m$ with $N$ such that the growth of the upper bound for $||\widehat{f}^{(m)}||_H^2$ can be controlled with a high probability. Observe that Theorem~\ref{theorem:hanson_wright} requires centered random variables $Z_1, \ldots ,Z_N$. 
\begin{lemma}\label{lemma:upper_bound_hilbert_space_norm}
Decompose $Y = \mu(\mathbf{\widetilde{X}}) + \varepsilon(\mathbf{\widetilde{X}})$, where $\mu(\mathbf{\widetilde{X}}) = \mathbb{E}\left[Y | \mathbf{\widetilde{X}}\right]$ and $\varepsilon(\mathbf{\widetilde{X}}) = Y - \mathbb{E}\left[Y | \mathbf{\widetilde{X}}\right]$. Assume that $||\mu||_\infty < \mu_{max}$ and the Orlicz norm and variance of the conditional distribution of $\varepsilon(\mathbf{\widetilde{X}})$ given some realization $\mathbf{\widetilde{x}}$ of $\mathbf{\widetilde{X}}$ is uniformly bounded by $s_{max}$ and $\sigma^2_{max}$, respectively. Let $\widehat{f}^{(m)}$ be the boosting estimate after $m$ boosting steps and $\Lambda, D, U \in \mathbb{R}^{N \times N}$ as in Lemma~\ref{lemma:expression_h_norm}. 
We can upper bound
\begin{align*}
\mathbb{P}\left(||\widehat{f}^{(m)}||_H^2 > 1 + 2 N (\mu_{max}^2 + \sigma_{max}^2) ||M_m^{1/2}(\mathbf{\widetilde{x}}^N)||^2_F |\mathbf{\widetilde{X}}^N = \mathbf{\widetilde{x}}^N\right) \\ \leq 2\exp\left(-C \min\left\{\frac{1}{4s_{max}^4  ||M_m||_F^2}, \frac{1}{2s_{max}^2 ||M_m||}\right\}\right).    
\end{align*}

where 
$$
M_m(\mathbf{\widetilde{x}}^N) := M_m := \frac{1}{N}U(I - (I - D)^m)^2 \Lambda^{-1} U^T.
$$
\end{lemma}

\begin{proof}
By Lemma~\ref{lemma:expression_h_norm} it holds $$||\widehat{f}^{(m)}||_H^2 = \frac{1}{N} ({y^N})^T U (I - (I - D)^m)^2 \Lambda^{-1} U^T {y^N} = \left(y^N\right)^\top M_m(\widetilde{\mathbf{x}}^N) y^N.$$
We emphasize that $G$ and thus $D$, $U$ and $S$ are functions of $\widetilde{\mathbf{X}}^N$. 
We calculate for $\mathbf{\mu}^N = \left(\mu(\widetilde{\mathbf{x}}_1), \ldots, \mu(\widetilde{\mathbf{x}}_N)\right) \in \mathbb{R}^N$ and $\mathbf{\varepsilon}^N = \left(\varepsilon_1(\widetilde{\mathbf{x}}_1), \ldots, \varepsilon_N(\widetilde{\mathbf{x}}_N)\right) \in \mathbb{R}^N$:
\begin{align*}
  &\mathbb{P}\left( ||\widehat{f}^{(m)}||_H^2 - 2 N (\mu_{max}^2 + \sigma_{max}^2) ||M_m^{1/2}(\widetilde{\mathbf{X}}^N)||^2_F > 1 | \widetilde{\mathbf{X}}^N = \widetilde{\mathbf{x}}^N \right) \\
  &= \mathbb{P}\left( ||M_m^{1/2}(\widetilde{\mathbf{X}}^N) Y^N||^2 - 2 N(\mu_{max}^2 + \sigma_{max}^2) ||M_m^{1/2}(\widetilde{\mathbf{X}}^N)||^2_F > 1 | \widetilde{\mathbf{X}}^N = \widetilde{\mathbf{x}}^N\right) \\
  &\leq \mathbb{P}\biggl( 2||M_m^{1/2}(\widetilde{\mathbf{X}}^N) \mathbf{\mu}^N||^2 + 2||M_m^{1/2}(\widetilde{\mathbf{X}}^N) \mathbf{\varepsilon}^N||^2   \\
  &\qquad - 2 N (\mu_{max}^2 + \sigma_{max}^2) ||M_m^{1/2}(\widetilde{\mathbf{X}}^N)||^2_F > 1 | \widetilde{\mathbf{X}}^N = \widetilde{\mathbf{x}}^N\biggr) 
  \end{align*}
In the third line we used the decomposition $y^N = \mu^N + \varepsilon^N$ and the inequality $||a + b||^2 \leq 2||a||^2 + 2 ||b||^2$.
In the following we upper-bound the terms $||M_m^{1/2}(\mathbf{\widetilde{X}}^N) \mu^N ||$ and $-N ||M_m^{1/2}(\widetilde{\mathbf{X}}^N)||^2_F \sigma_{max}^2$. For the first term observe that
\begin{align*}
||M_m^{1/2}(\mathbf{\widetilde{X}}^N) \mu^N||^2 &= tr\left((\left(\mu^N\right)^\top M_m(\mathbf{\widetilde{X}}^N) \mu^N\right) \\
&= tr\left(M_m(\mathbf{\widetilde{X}}^N)  \mu^N \left(\mu^N\right)^\top \right) \\
&\leq tr\left(M_m(\mathbf{\widetilde{X}}^N)\right) tr\left(\mu^N \left(\mu^N\right)^\top \right) \\
&\leq N ||M_m^{1/2}\left(\mathbf{\widetilde{X}}^N\right)||^2_F \mu_{max}^2.    
\end{align*}
For the latter term using that $\mathbf{E}\left[\varepsilon_k^2 | \widetilde{\mathbf{X}}^N \right] \leq \sigma^2_{max}, k = 1, \ldots, N $ it holds analogously
\begin{align*}    
\mathbb{E}\left[||M_m^{1/2}(\widetilde{\mathbf{X}}^N)
\varepsilon^N||^2 | \widetilde{\mathbf{X}}^N \right] &= \mathbb{E}\left[tr\left(\left(\varepsilon^N\right)^\top M_m(\widetilde{\mathbf{X}}^N)
\varepsilon^N \right) | \widetilde{\mathbf{X}}^N \right] \\
&\leq \mathbb{E}\left[tr\left(M_m(\widetilde{\mathbf{X}}^N)\right)tr\left(\varepsilon^N \left(\varepsilon^N\right)^\top 
 \right) | \widetilde{\mathbf{X}}^N \right] \\
&\leq tr\left(M_m(\widetilde{\mathbf{X}}^N)\right)\mathbb{E}\left[
\left(\varepsilon^N\right)^\top \varepsilon^N  | \widetilde{\mathbf{X}}^N \right] \\
&\leq  N ||M_m^{1/2}(\widetilde{\mathbf{X}}^N)||^2_F \sigma_{max}^2
\end{align*}
and hence $-N ||M_m^{1/2}(\widetilde{\mathbf{X}}^N)||^2_F \sigma_{max}^2 \leq -\mathbb{E}\left[||M_m^{1/2}(\widetilde{\mathbf{X}}^N)
\varepsilon^N||^2 | \widetilde{\mathbf{X}}^N \right]$.
Using these results and plugging in the condition $\widetilde{\mathbf{X}}^N = \widetilde{\mathbf{x}}^N$ we can upper-bound
\begin{align*}
&\mathbb{P}\left( 2||M_m^{1/2} \mu^N||^2 + 2||M_m^{1/2} \varepsilon^N||^2  - 2 N (\mu_{max}^2 + \sigma_{max}^2) ||M_m^{1/2}(\widetilde{\mathbf{X}}^N)||^2_F > 1 | \widetilde{\mathbf{X}}^N = \widetilde{\mathbf{x}}^N \right) \\
  &\leq \mathbb{P}\left( ||M_m^{1/2} \varepsilon^N||^2  - \mathbb{E}\left(||M_m^{1/2}(\widetilde{\mathbf{X}}^N) \varepsilon^N||^2 | \widetilde{\mathbf{X}}^N \right) > \frac{1}{2} | \widetilde{\mathbf{X}}^N = \widetilde{\mathbf{x}}^N \right) \\
  &= \mathbb{P}\left( ||M_m^{1/2} \varepsilon^N||^2  - \mathbb{E}\left(||M_m^{1/2} \varepsilon^N||^2  \right) > \frac{1}{2} | \widetilde{\mathbf{X}}^N = \widetilde{\mathbf{x}}^N\right) \\
  &\leq 2\exp\left(-C \min\left\{\frac{1}{4s_{max}^4  ||M_m||_F^2}, \frac{1}{2s_{max}^2 ||M_m||}\right\}\right)
\end{align*}
by Theorem~\ref{theorem:hanson_wright} (Hanson-Wright-inequality) and the fact that $\varepsilon^N | \widetilde{\mathbf{X}}^N = \widetilde{\mathbf{x}}^N$ is a centered, sub-Gaussian random vector with independent components with Orlicz norm bounded by $s_{max}$. 
\end{proof}
We now show that if we choose $m = m(N) 
 = N^{\frac{1}{4}\frac{C_u + C_d + 1/2}{C_d + 1}}$ then the growth of $||\widehat{f}^{(m)}||_H$ with $N$ is of lower order as $N^{1/4}$ with a probability going to $1$ if $\widetilde{\mathbf{x}}^N \in \mathcal{B}_N$.
\begin{lemma}[Upper bound $||\widehat{f}^{(m_{stop})}||_H$]\label{lemma:boosting_estimate_in_scaled_spaces}
    Under Assumption~\ref{ass:decay_eigenvalues} 
    and for \newline$m_{stop} = m(N) = N^{\frac{1}{4}\frac{C_u + C_d + 1/2}{C_d + 1}}$
    there exists a $\delta > 0$ and
    a $h(N) \in o(N^{1/4 - \delta})$ so that
    \begin{equation}\label{eq:convergence_lemma_5}
    \mathbb{P}\left(\widehat{f}^{(m_{stop})} \notin h(N) \mathcal{F}_N \cap \{\mathbf{\widetilde{X}}^N \in \mathcal{B}_N \}\right) \rightarrow 0    
    \end{equation}
    for $N \rightarrow \infty$.
\end{lemma}
\begin{proof}
As outlined in relation~\eqref{eq:structure_function_estimate}, by the representation theorem it holds that $\widehat{f}^{(m)} \in h(N) \mathcal{F}_N$ for some $h(N) \in \mathbb{R}$. Thus we need to show that $||\widehat{f}^{(m_{stop})}||_H \leq h(N)$. We prove the statement by showing the convergence
$$
\mathbb{P}\left(||\widehat{f}^{(m_{stop})}||^2_H > h(N)^2 | \mathbf{\widetilde{X}}^N = \mathbf{\widetilde{x}}^N\right) \rightarrow 0
$$
uniformly for any $\mathbf{\widetilde{x}}^N \in \mathcal{B}_N$ and some $h(N) \in o(N^{1/4 - \delta})$. The statement then follows by integrating out with respect to $\mathbf{\widetilde{x}}^N$. Applying Lemma~\ref{lemma:upper_bound_hilbert_space_norm}, 
we need to prove the following two statements.
    \begin{enumerate}
        \item $N||M_m^{1/2}||_F^2 \in o\left(N^{1/2 - 2\delta}\right)$. This implies, $||M_m||_F^2 \leq ||M_m^{1/2}||^4_F \in o(1)$. 
        \item $||M_m|| \in o(1)$.
    \end{enumerate}
It is emphasized that $M_m$ is a function of the random sample $\mathbf{\widetilde{x}}^N$. \newline
The statement is proven in Lemma~\ref{lemma:convergence_M_m}.
\end{proof}
\begin{lemma}\label{lemma:convergence_M_m}
Under Assumption~\ref{ass:decay_eigenvalues} it holds for
$$
M_m(\mathbf{\widetilde{x}}^N) := M_m := \frac{1}{N}U(I - (I - D)^m)^2 \Lambda^{-1} U^T
$$
that there exists a $\delta > 0$ so that the following statements hold for $m = m(N) = N^{\frac{1}{4}\frac{C_u + C_d + 1/2}{C_d + 1}}$ uniformly in $\mathbf{\widetilde{x}}^N \in \mathcal{B}_N$, where $\mathcal{B}_N$ is defined in Assumption~\ref{ass:decay_eigenvalues}.
\begin{enumerate}
    \item $
    N^{1/2 + 2\delta}||M_m^{1/2}||^2_F = \frac{1}{N^{1/2 - 2\delta}}tr\left((I - (I - D)^m)^2 \Lambda^{-1}\right)\rightarrow 0,
    $
    \item $||M_m||_F^2 \rightarrow 0$, and
    \item $||M_m|| \rightarrow 0$.
\end{enumerate}
\end{lemma}
\begin{proof}
\textbf{1.:} Uniformly in $\mathbf{\widetilde{x}}^N \in \mathcal{B}_N$ it holds that: 
\begin{align*}
N^{1/2 + 2\delta}||M_m^{1/2}||_F^2 &= N^{1/2 + 2\delta} tr(M_m)  \\
&= \frac{1}{N^{1/2 - 2\delta}}\sum_{k=1}^N (1 - (1- d_k)^m)^2 \widehat{\mu_k}^{-1} \\
&= \frac{1}{N^{1/2 - 2\delta}}\sum_{k=1}^N \left(1 - \left(1- \frac{\widehat{\mu_k}}{\widehat{\mu_k} + \lambda}\right)^m\right)^2 \widehat{\mu_k}^{-1} \\
&\stackrel{(i)}{\leq} \frac{1}{N^{1/2 - 2\delta}}\sum_{k=1}^N \min\left\{1, m^2 \left(\frac{\widehat{\mu_k}}{\widehat{\mu_k} + \lambda}\right)^2\right\} \widehat{\mu_k}^{-1} \\
&\leq \frac{1}{\lambda^{2}N^{1/2 - 2\delta}}\sum_{k=1}^N \min\left\{\lambda^{2}\widehat{\mu_k}^{-1} , m^2 \widehat{\mu_k}\right\} \\
&\leq \frac{1}{\lambda^{2} N^{1/2 - 2\delta}}\left(\sum_{k=1}^{\lfloor K_0 \rfloor} \min\left\{\lambda^{2}\widehat{\mu_k}^{-1}, m^2 \widehat{\mu_k}\right\} + \sum_{k=\lceil K_0 \rceil}^N \min\left\{\lambda^2\widehat{\mu_k}^{-1}, m^2 \widehat{\mu_k}\right\}\right)\\
&\leq \frac{1}{\lambda^2 N^{1/2 - 2\delta}} \left(K_0\widehat{\mu_{K_0}}^{-1} \lambda^{2} + \sum_{k=\lceil K_0 \rceil}^N m^2 \widehat{\mu_k} \right),
\end{align*}
which holds for any $K_0 \in \mathbb{N}$. The inequality $(i)$ is shown in Lemma~\ref{lemma:shrinkage}. The last inequality is due to the fact that $\widehat{\mu}^{-1}_k$ is monotonically increasing.
We choose $K_0 = K_0(N) = \lfloor\frac{1}{2C_d + 1}\ln(N)\rfloor$. Uniformly in $\mathbf{\widetilde{x}}^N \in \mathcal{B}_N$ it holds by Assumption~\ref{ass:decay_eigenvalues} for a small $\delta > 0$

$$\frac{K_0\widehat{\mu_{K_0}}^{-1}}{N^{1/2 - 2\delta}} \leq \frac{K_0 \exp(C_d K_0)}{N^{1/2 - 2\delta}}  \rightarrow 0.$$
For the latter part observe that it holds for any $\mathbf{\widetilde{x}}^N \in \mathcal{B}_N$ and by Assumption~\ref{ass:decay_eigenvalues}
\begin{align*}
\frac{1}{N^{1/2 - 2\delta}}\sum_{k=\lceil K_0 \rceil}^N m^2 \widehat{\mu_k} &\leq \frac{m^2}{N^{1/2 - 2\delta}} \sum_{k=\lceil K_0 \rceil}^N \exp(-C_u k)\\
&\leq \frac{m^2}{N^{1/2 - 2\delta}}   \int_{K_0 }^\infty  \exp(-C_u (z-1)) dz \\
&= \frac{m^2 \exp(C_u)}{N^{1/2 - 2\delta} C_u}   \exp(-C_u K_0) \\
&= \frac{m^2 \exp(C_u)}{N^{1/2 - 2\delta} C_u}   \exp(-C_u (\underbrace{K_0 + 1}_{\geq \frac{1}{2C_d + 1}\ln(N)} - 1)) \\
&\leq \frac{m^2 \exp(2C_u)}{N^{1/2 - 2\delta} C_u}  N^{-\frac{C_u}{2C_d + 1}}\\
&= \frac{m^2 \exp(2C_u)}{C_u}  N^{-\frac{C_u + C_d + 1/2}{2C_d + 1}} N^{2\delta}.
\end{align*}
Observe that for $m=m(N) = N^{\frac{1}{4}\frac{C_u + C_d + 1/2}{C_d + 1}}$ where the constants are chosen independently of $\mathbf{\widetilde{x}}^N$, it holds 
$$
m^2 N^{-\frac{C_u + C_d + 1/2}{2C_d + 1}} = N^{\frac{C_u + C_d + 1/2}{2C_d + 2} - \frac{C_u + C_d + 1/2}{2C_d + 1}} = N ^{-\xi},
$$
where $\xi := \frac{C_u + C_d + 1/2}{2C_d + 1} - \frac{C_u + C_d + 1/2}{2C_d + 2} > 0$. For $\delta < \frac{\xi}{2}$ it holds that $\frac{1}{N^{1/2 - 2\delta}}\sum_{k=\lceil K_0 \rceil}^N m^2 \widehat{\mu_k} \rightarrow 0$. \newline
\textbf{2.} follows by $||M_m||_F^2 \leq ||M_m^{1/2}||^4_F \rightarrow 0$. \newline
\textbf{3.} follows by 
\begin{align*}
||M_m|| &= \frac{1}{N}\max_{k=1, \ldots, N} \left(1 - \left(1-\frac{\widehat{\mu}_k}{\widehat{\mu}_k + \lambda}\right)^m\right)^2 \widehat{\mu}_k^{-1} \\  &\stackrel{(i)}{\leq}
\frac{1}{N}\max_{k=1, \ldots, N} \min\left\{1, m^2 \left(\frac{\widehat{\mu_k}}{\widehat{\mu_k} + \lambda}\right)^2\right\} \widehat{\mu_k}^{-1}  \\
&\leq \frac{1}{\lambda^2  N } \max_{k=1, \ldots, N} \min\left\{\widehat{\mu_k}^{-1}, m^2 \widehat{\mu_k}\right\} \leq \frac{m^2 \widehat{\mu}_1}{\lambda^2 N } \rightarrow 0,
\end{align*}
where inequality $(i)$ follows again from Lemma~\ref{lemma:shrinkage} as $\widehat{\mu}_1 \leq 1$.
\end{proof}
The following Lemma immediately follows from the proof of Lemma~\ref{lemma:convergence_M_m}.
\begin{lemma}\label{lemma:decay_evs_consistency_boosting}
    Under Assumption~\ref{ass:decay_eigenvalues} it holds for $m = m(N) = N^{\frac{1}{4}\frac{C_u + C_d + 1/2}{C_d + 1}}$ on $\mathcal{B}_N$, that
    $$
    \frac{1}{\sqrt{N}}\sum_{\ell=1}^N \left(1 - \left(1- \frac{\widehat{\mu_\ell}}{\widehat{\mu_\ell} + \lambda}\right)^m\right)^2  \rightarrow 0.
    $$
\end{lemma}
\begin{lemma}\label{lemma:shrinkage}
It holds for $0 \leq \widehat{\mu_k} \leq 1$, that 
$$
1- \left(1 - \widehat{\mu_k}\right)^m \leq 1 - \max\left\{0, 1 - m \widehat{\mu_k}\right\} = \min\left\{1,  m \widehat{\mu_k}\right\}
$$
\end{lemma}

\begin{proof}
It is equivalent to show that 
$$
(1- \widehat{\mu_k})^m \geq \max\{0, 1-m\widehat{\mu_k}\}.
$$
The l.h.s. and the r.h.s. are equal for $\widehat{\mu_k} = 0$. On the interval $[0, \frac{1}{m})$ it holds that 
$$
\frac{\partial(1- \widehat{\mu_k})^m}{\partial\widehat{\mu_k}} = -m (1-\widehat{\mu_k})^{m-1} \geq -m = \frac{\partial(\max\{0, 1-m\widehat{\mu_k}\}}{\partial\widehat{\mu_k}}.
$$
Hence, 
$$
(1- \widehat{\mu_k})^m \geq \max\{0, 1-m\widehat{\mu_k}\} \text{ for } \widehat{\mu_k} \in [0, \frac{1}{m}).
$$
Clearly, for $\mu_k \in [\frac{1}{m}, 1]$ 
$$
(1- \widehat{\mu_k})^m \geq \max\{0, 1-m\widehat{\mu_k}\}.
$$
\end{proof}

%% file: Y_Appendix/B2_covering_numbers.tex
Lemma~\ref{lemma:boosting_estimate_in_scaled_spaces} has shown that for  $m_{stop} = N^{\frac{1}{4}\frac{C_u + C_d + 1/2}{C_d + 1}}$, it holds $\widehat{f}^{(m_{stop})} \in h(N) \mathcal{F}_N$ for some $h(N) \in o(N^{1/4})$ with probability going to $1$ for $N \rightarrow \infty$. In this section we use the covering numbers from empirical process theory to show the convergence of the inner product
\begin{equation}\label{eq:inner_product_convergence_appendix}
|(P- P_N) Y\widehat{f}^{(m_{stop})}|.
\end{equation}
The covering numbers measure the complexity of a function class. 
\begin{definition}
For a function class $\mathcal{F}$ and a semi-metric $d$ on $\mathcal{F}$ the covering number 
$\mathcal{N}(\varepsilon, \mathcal{F}, d)$
is the minimal size of a subset $S \subset \mathcal{F}$, such that for every $f \in \mathcal{F}$ there is an $s \in S$ so that $d(f,s) < \varepsilon$. More precisely,
$$
\mathcal{N}(\varepsilon, \mathcal{F}, d) = \min_{\{S \subset \mathcal{F}|\forall f \in \mathcal{F} \, \exists s \in S: d(f, s) < \varepsilon \}} |S|.
$$
In this work, we choose $d(f,g) = ||f - g||_\infty$ and thus write $\mathcal{N}\left(u, \mathcal{F}, ||\cdot||_{\infty}\right) = \mathcal{N}\left(u, \mathcal{F}\right)$.
For a suitable $C_0$ \citep[chosen as in Dudley's Theorem, see Theorem 8.4 of][]{empProcessesCAM}
not depending on $\mathcal{F}$ we define the covering number entropy integral by
$$
\mathcal{J}(z, \mathcal{F}) := C_0 z\int_0^1 \sqrt{\log{\mathcal{N}\left(\frac{uz}{2}, \mathcal{F}\right)}}  du .
$$
\end{definition}
For a constant $C > 0$, let $\mathcal{H} = \{C h | h \in \mathcal{G}\}$ be a scaled version of some function class $\mathcal{G}$. The following remark shows that the entropy integral $\mathcal{J}(z, \mathcal{H})$ of $\mathcal{H}$ can be upper bounded by an expression depending on $C$ and the entropy integral $\mathcal{J}(z, \mathcal{G})$ of $\mathcal{G}$.
\begin{remark}\label{remark:log_entropy_integral}
For $C > 0$ the identity
$$
\mathcal{N}(C z, C \mathcal{G}, ||\cdot ||) \leq \mathcal{N}(z, \mathcal{G}, ||\cdot ||)
$$
holds and thus
$
\mathcal{J}(Cz, C\mathcal{G}) \leq C \mathcal{J}(z, \mathcal{G}),
$
where $C\mathcal{G} = \left\{ C g | g \in \mathcal{G}\right\}$.
This upper bound is not optimal \citep[see][] {cucker2002mathematical} but sufficient for our purposes.
\end{remark}
Eventually, we will show in Corollary~\ref{corollary:covering_numbers_grow_h} that if $\widehat{f}^{(m_{stop})}$ is in the ball of radius $h(N) \in o(N^{1/4})$, then this growth rate $h(N)$ is slow enough to ensure the convergence of \eqref{eq:inner_product_convergence_appendix}. We rely on the following theorem.

\begin{theorem}[{{\citet[Theorem 3.2]
{empProcessesCAM}}}]
\label{theorem:deviation_inner_product}
    Let $K = \sup_{f \in \mathcal{F}} ||f||_\infty$ and assume that $Y$ is sub-Gaussian with Orlicz norm smaller than $s$. Then for $t,N$ such that \footnote{There is a typo in \cite{empProcessesCAM}, where it says $J_0 (K\sigma, \mathcal{F})$ instead of $J_\infty (K\sigma, \mathcal{F})$.}\newline 
$\sqrt{\frac{2t}{N}} + \frac{t}{N} \leq 1$
it holds with probability $1 - 8\exp(-t)$
\begin{equation}\label{eq:theorem2}
\sup_{f \in \mathcal{F}} | (P_N -P) Yf | / C 
\leq \frac{2\mathcal{J}(K s, \mathcal{F})  + K s\sqrt{t}}{  \sqrt{N}}.
\end{equation}
\end{theorem}
\begin{corollary}\label{corollary:covering_numbers_grow_h} Let $f_1, f_2, \ldots$ be a sequence in $H$ such that
    $$f_N \in h(N) \mathcal{F}_N,$$
    where $h(N) \in o(N^{1/4})$. If $\mathcal{J}\left(z, \mathcal{F}_N\right) \leq \mathcal{J}\left(z, B_1\right) = C_0 z \int_0^1 \sqrt{\log\left( \mathcal{N}(\frac{uz}{2}, B_1)\right)} 
    < \infty$ for all $z > 0$ as in Assumption~\ref{ass:covering_number_hilbert_space}, then
    $$
    |(P_N - P) Yf_N| \leq \xi.
    $$
    converges to $0$ in probability.
\end{corollary}
\begin{proof}
    For $h(N) \mathcal{F}_N$ it holds that
    $$
    K := \sup_{f \in h(N) \mathcal{F}_N} ||f||_\infty \leq \sup_{f \in h(N)B_1} ||f||_\infty \leq B h(N).
    $$
    Recalling the definition of $\mathcal{J}(u, \mathcal{F})$ and applying Remark~\ref{remark:log_entropy_integral}, we obtain
    $$
    \mathcal{J}(Ku, h(N) \mathcal{F}_N) \leq \mathcal{J}(B h(N)u, h(N) B_1) \leq h(N) \mathcal{J}(Bu, B_1) \, \forall u \in \mathbb{R}_+.
    $$

    As the Orlicz norm fulfills the triangle inequality and as the Orlicz norm of the bounded random variable $\mu(\mathbf{\widetilde{X}})$ is finite, the Orlicz norm of $Y = \mu(\mathbf{\widetilde{X}})  + \varepsilon$, denoted by $s$, is bounded and $Y$ is thus sub-Gaussian.
    We now apply Theorem~\ref{theorem:deviation_inner_product} and set $t = N^{1/2}$ (the condition $\sqrt{\frac{2t}{N}} + \frac{t}{N} \leq 1$ is then fulfilled for $N > \frac{1}{2}\left(7 + 3\sqrt{5}\right)$).

    It holds with probability $1 - 8\exp(-N^{1/2})$
    \begin{align*}
    |(P_N - P) Yf_N| &\leq \sup_{f \in h(N)\mathcal{F}_N} |(P_N - P) Yf| \\
    &\leq \frac{\mathcal{J}( B h(N) s, h(N)\mathcal{F}_N)  + B h(N)  s_{max}N^{1/4}}{  \sqrt{N}} \\
    &\leq \frac{h(N)\mathcal{J}(B s, \mathcal{F}_N)  + B h(N)  s_{max}N^{1/4}}{  \sqrt{N}} \\
    &\leq \frac{h(N)\mathcal{J}(B s, B_1)  + B h(N)  s_{max}N^{1/4}}{  \sqrt{N}}.
    \end{align*}
    $\mathcal{J}(B s, B_1)$ is finite and constant in $N$ by Assumption~\ref{ass:covering_number_hilbert_space}. Let $\xi > 0$ be arbitrary. As $h(N) \in o(N^{1/4})$, there exists $N^0$ so that  
    $$
    \frac{h(N)\mathcal{J}(B s, B_1)}{\sqrt{N}} \leq \frac{\xi}{2} \, \forall N > N^0.
    $$
    For the second term observe that as $h(N) \in o(N^{1/4})$
    $$\frac{B h(N)  s_{max} N^{1/4}}{  \sqrt{N}} = \frac{B h(N)  s_{max} }{  N^{1/4}} \leq \frac{\xi}{2}$$
    for all $N > N^1$ for some $N^1 \in \mathbb{N}$. Thus for $N > \max\{N^0, N^1\}$
    $$|(P_N - P) Yf_N| \leq \xi$$
    with probability $1 - 8\exp(-8N^{1/2}) \rightarrow 1$. This proves the convergence in probability.
\end{proof}

%% file: Y_Appendix/B3_rademacher_complexity.tex
In this section we use concept of the Rademacher complexity to show the convergence of
\begin{equation*}
\left|(P- P_N)\left(\widehat{f}^{(m_{stop})}\right)^2\right|.
\end{equation*}
It is again based on Lemma~\ref{lemma:boosting_estimate_in_scaled_spaces}, which ensures that for  $m_{stop} = N^{\frac{1}{4}\frac{C_u + C_d + 1/2}{C_d + 1}}$, it holds \newline$\widehat{f}^{(m_{stop})} \in h(N) \mathcal{F}_N$ for some $h(N) \in o(N^{1/4 - \delta})$ for some $\delta > 0$ with probability going to $1$ for $N \rightarrow \infty$.
\begin{definition}[Rademacher complexity]
Let $\mathcal{F}$ be a function class on $\mathbf{X}$. Let $\sigma_1, \ldots, \sigma_N$ be i.i.d. realizations of Rademacher random variables, which are independent of $\mathbf{X}$. Further let $\mathbf{x}_1, \ldots, \mathbf{x}_N$ be $i.i.d$ realizations of $\mathbf{X}$. We define the Rademacher complexity $R_N(\mathcal{F})$ by

$$
R_N(\mathcal{F}) = \frac{1}{N}\mathbb{E}_{\mathbf{X}, \sigma}\left[\sup_{f \in \mathcal{F}} \left| \sum_{\ell = 1}^N \sigma_\ell f(\mathbf{x}_\ell) \right| \right].
$$
Note that $R_N(\mathcal{F})$ is deterministic.
Given fixed observations $\mathbf{x}_1, \ldots, \mathbf{x}_N$, the empirical Rademacher complexity is given by
$$
\widehat{R}_N(\mathcal{F}|\mathbf{x}_1, \ldots, \mathbf{x}_N) = \widehat{R}_N(\mathcal{F}) = \frac{1}{N}\mathbb{E}_{\sigma}\left[\sup_{f \in \mathcal{F}} \left|\sum_{\ell = 1}^N \sigma_\ell f(\mathbf{x}_\ell) \right| \right].
$$
\end{definition}
The Rademacher complexity is a tool to upper bound $\sup_{f \in \mathcal{F}}\left|(P_N - P)  f \right|$. 
\begin{theorem}\label{theorem:uniform_convergence_rademacher}\cite[Theorem 4.10]{wainwright_hds} Let $\mathcal{F}$ be uniformly bounded with constant $B$. Then it holds for any $N \in \mathbb{N}$ and with probability $1 - \exp(-\frac{\varepsilon^2 N}{2B^2})$, that

$$
\sup_{f \in \mathcal{F}}\left|(P_N - P)  f \right| 
\leq 2 R_N(\mathcal{F}) + \varepsilon.
$$
    
\end{theorem}
The Rademacher complexity is linked to its empirical counterpart with high probability. 
\begin{theorem}
[{{\citet[Theorem 11]{bartlett2002rademacher}}}]\label{prop:empirical_rademacher_convergence}
    Let $\mathcal{F}$ be uniformly bounded by $B$. Then it holds with probability $1 - 2\exp\left(-\frac{\varepsilon^2 N}{B^2}\right)$
    $$
    \left|\widehat{R}_N(\mathcal{F}) - R_N(\mathcal{F}) \right| < \varepsilon.
    $$
\end{theorem}
We collect some helpful relationships for the Rademacher complexity. 
\begin{theorem}[{{\citet[Theorem 12]{bartlett2002rademacher}}}]
\label{theorem:rademacher_structural}
    Let $\mathcal{F}, \mathcal{F}_1, \mathcal{F}_2, \ldots, \mathcal{F}_p$ be function classes and let $\mathcal{F}$ be uniformly bounded by $B$. Then 
    \begin{enumerate}
        \item for $c \in \mathbb{R}$: $R_N(c\mathcal{F}) = |c|R_N(\mathcal{F})$,
        \item $R_N\left(\sum_{j=1}^p \mathcal{F}_j \right) \leq \sum_{j=1}^p R_N\left( \mathcal{F}_j \right)$,
        \item for $\mathcal{F}^2 := \{|f|^2: f \in \mathcal{F}\}$, it follows  
        $
        R_N(\mathcal{F}^2) \leq 4 B R_N(\mathcal{F}).
        $
    \end{enumerate}
\end{theorem}
The Rademacher complexity can be upper bounded for kernel functions. 
\begin{theorem}[{{\citet[Lemma 22]{bartlett2002rademacher}}}]
\label{theorem:rademacher_kernel}
    The empirical Rademacher complexity of $\mathcal{F}_N$ is upper bounded by
    $$
    \widehat{R}_N(\mathcal{F}_N) \leq \left(\frac{2}{N}\sum_{k=1}^N \widehat{\mu_k}\right)^{\frac{1}{2}},
    $$
    while the population Rademacher complexity can be upper bounded by
    $$
        R_N(\mathcal{F}_N) \leq \left(\frac{2}{N}\sum_{k=1}^N \mu_k\right)^{\frac{1}{2}}.
    $$
\end{theorem}
Note that if the sequence $\mu_1, \mu_2, \ldots$ is summable, then $R_N(\mathcal{F}_N) \in O(N^{-1/2})$. For this case we connect the results above in a corollary.
\begin{corollary}\label{corollary:uniform_convergence_norm_grow_h}
    For any sequence $f_1, f_2, \ldots$ in $H$ for which
    $$f_N \in h(N) \mathcal{F}_N,$$
    where $h(N) \in o(N^{1/4 - \delta})$ for some $\delta > 0$ and $\widehat{R}_N(\mathcal{F}_N) \in O(N^{-1/2})$, it holds
    $$
    |(P_N - P) f_N^2| \overset{\mathbb{P}}{\rightarrow} 0 \text{ for } N \rightarrow \infty.
    $$
\end{corollary}
\begin{proof}
Let $\xi > 0$ be arbitrary and fixed. By Theorem~\ref{theorem:rademacher_structural} it holds that 
    $$
    R_N(h(N)\mathcal{F}_N) = h(N) R_N(\mathcal{F}_N),
    $$
    and as $\mathcal{F}_N$ is uniformly bounded by $B h(N)$ it holds by Theorem~\ref{theorem:rademacher_structural}
    $$
    R_N\left((h(N)\mathcal{F}_N)^2\right) = 4B h(N) R_N(\mathcal{F}_N).
    $$
    From Theorem~\ref{theorem:uniform_convergence_rademacher} we obtain that for any $\xi > 0$
    $$
    |(P_N - P) f_N^2| \leq \sup_{f \in h(N)\mathcal{F}_N}\left|(P_N - P)  f^2 \right| \leq 2 R_N\left((h(N)\mathcal{F}_N)^2\right) + \frac{\xi}{2} =  8 B h(N)^2 R_N(\mathcal{F}_N) + \frac{\xi}{3}
    $$
    with probability $1 - 2\exp\left(\frac{\xi^2 N}{4(Bh(N))^2}\right)$, which converges to $1$ as $h(N) \in o(N^{1/4})$ for $N \rightarrow \infty$.
    Similarly, as $\mathcal{F}_N$ is uniformly bounded by $B$ and by setting $\varepsilon = N^{-1/2  + 2\delta}$ in Theorem~\ref{prop:empirical_rademacher_convergence}, we observe that
    $$
    |\widehat{R}_N(\mathcal{F}_N) - R_N(\mathcal{F}_N)| \leq N^{-1/2  + 2\delta}
    $$
    with probability going to $1$ for any $\delta > 0$. We conclude that
    \begin{align*}
        |(P_N - P) f_N^2| &\leq \sup_{f \in h(N)\mathcal{F}_N}\left|(P_N - P)  f^2 \right| \\
        &\leq 8 B h(N)^2 R_N(\mathcal{F}_N) + \frac{\xi}{3} \\
        &\leq 8 B h(N)^2 |R_N(\mathcal{F}_N) - \widehat{R}_N(\mathcal{F}_N) | + 8 B h(N)^2 \widehat{R}_N(\mathcal{F}_N)  + \frac{\xi}{3} \\
        &\leq 8B h(N)^2 N^{-1/2 +2\delta} + 8B h(N)^2 \widehat{R}_N(\mathcal{F}_N) +  \frac{\xi}{3}
    \end{align*}
with probability going to $1$ for $N \rightarrow \infty$. As $h(N) \in o(N^{1/4})$, it holds $h(N)^2 \widehat{R}_N(\mathcal{F}_N) < \frac{\xi}{3}$ by Assumption~\ref{ass:decay_eigenvalues} and Theorem~\ref{theorem:rademacher_kernel} for $N$ chosen large enough. Similarly, $ h(N)^2 N^{-1/2  + 2\delta} < \frac{\xi}{3}$ for $N$ chosen large enough, as $h(N) \in o\left(N^{1/4 - \delta}\right)$. This proves the statement.
\end{proof}

%% file: Y_Appendix/C_fixed_design_convergence.tex
\begin{proof}[Proof of Lemma~\ref{lemma:fixed_design_convergence}]
Using
\begin{align*}
    ||f^0 - \widehat{f}^{(m)}||^2_{2,N} &= \frac{1}{N} ||f^0(\mathbf{\widetilde{x}}^N) - B^{(m)} y^N||^2_2 \\
    &= \frac{1}{N} ||f^0(\mathbf{\widetilde{x}}^N) - U \left( I - \left(I-D\right)^m\right) U^\top  y^N||^2_2 \\
    &= \frac{1}{N} ||f^0(\mathbf{\widetilde{x}}^N) - U \left( I - \left(I-D\right)^m\right) U^\top  \left(f^0(\mathbf{\widetilde{x}}^N) + \varepsilon^N\right)||^2_2  \\
    &\leq \frac{2}{N} ||f^0(\mathbf{\widetilde{x}}^N) - U \left( I - \left(I-D\right)^m\right) U^\top  f^0(\mathbf{\widetilde{x}}^N)||^2_2  \\ 
    & \quad + \frac{2}{N} ||U \left( I - \left(I-D\right)^m\right) U^\top  \varepsilon^N)||^2_2 \\
    &\leq \underbrace{\frac{2}{N} ||U \left(I-D\right)^m U^\top  f^0(\mathbf{\widetilde{x}}^N)||_2^2}_{\rm I}  + \underbrace{\frac{2}{N} ||U \left( I - \left(I-D\right)^m\right) U^\top \varepsilon^N ||^2_2}_{\rm II},
\end{align*}
where the first inequality holds due to $||a + b||^2 \leq 2||a||^2 + ||b||^2$, we show the convergence of $\rm I$ and $\rm II$ to $0$ in probability for $N \rightarrow \infty$. Recall that $S = U D U^T$ with $U$ containing the orthonormal eigenvalues of $S$ and $D$ being a diagonal matrix with diagonal entries $D_{\ell \ell} = d_\ell = \frac{\widehat{\mu}_\ell}{\widehat{\mu}_\ell + \lambda}$, where $\widehat{\mu}_\ell, \ell = 1, \ldots, N$ are the eigenvalues of $G$. \newline
\textbf{Convergence of $\rm I$:}
\begin{align*}
    \frac{2}{N} ||U \left(I-D\right)^m U^\top  f^0(\mathbf{\widetilde{x}}^N)||^2_2 &= \frac{2}{N} f^0(\mathbf{\widetilde{x}}^N)^\top U  \left(I-D\right)^m U^\top U \left(I-D\right)^m U^\top  f^0(\mathbf{\widetilde{x}}^N) \\
    &= \frac{2}{N} \sum_{\ell = 1}^N (1 - d_\ell)^{2m} \left( U^\top f^0(\mathbf{\widetilde{x}}^N) \right)^2_\ell
\end{align*}
As $G$ has full rank, there exists a $\beta \in \mathbb{R}^N$ such that $f^0(\mathbf{\widetilde{x}^N}) = \sqrt{N} G \beta$. Define $\widetilde{f} := \frac{1}{\sqrt{N}}\sum_{\ell = 1}^N \beta_\ell K(\cdot, \mathbf{\widetilde{x}}_\ell)$ for which holds $\widetilde{f}(\mathbf{\widetilde{x}}^N) = f^0(\mathbf{\widetilde{x}}^N)$. By the representation theorem it further holds $\beta^\top G \beta = ||\widetilde{f}||^2_H \leq ||f^0||^2_H = R^2$. Let $D_{\widehat{\mu}} \in \mathbb{R}^{N \times N}$ be the diagonal matrix with diagonal entries $\widehat{\mu}_1, \ldots, \widehat{\mu}_N$. Then,
\begin{align*}
    \frac{2}{N} \sum_{\ell = 1}^N (1 - d_\ell)^{2m} \left( U^\top f^0(\mathbf{\widetilde{x}}^N) \right)^2_\ell &\leq \frac{2}{N} \sum_{\ell = 1}^N \frac{\left( U^\top f^0(\mathbf{\widetilde{x}}^N) \right)^2_\ell}{2 e \, m \, d_\ell} \\
    &= \frac{1}{N} \sum_{\ell = 1}^N \frac{\left( U^\top \sqrt{N} G \beta \right)^2_\ell}{e \, m \, d_\ell} \\
    &= \sum_{\ell = 1}^N \frac{\left( U^\top  G \beta \right)^2_\ell}{e \, m \, d_\ell} \\
    &\leq (1+ \lambda) \sum_{\ell = 1}^N \frac{\left( U^\top   G \beta \right)^2_\ell}{e \, m \, \widehat{\mu}_\ell} \\
    &=  \frac{1+ \lambda}{e \, m} tr\left(D_{\widehat{\mu}}^{-1} U^\top  G  \beta \beta^\top G U\right) \\
    &=  \frac{1+ \lambda}{e \, m} tr\left(\underbrace{U D_{\widehat{\mu}}^{-1} U^\top}_{G^{-1}}  G  \beta \beta^\top G \right) \\
    &=  \frac{1+ \lambda}{e \, m}  \beta^\top   G \beta =  \frac{1+ \lambda}{e \, m} ||\widetilde{f}||^2_H \leq  \frac{1+ \lambda}{e \, m} R^2,
\end{align*}
which goes to $0$ as $m(N) \rightarrow \infty$ for $N \rightarrow \infty$. In the first inequality we have used the fact that $(1-x)^{2m} \leq \exp(-x)^{2m} = \exp(-2mx) \leq \frac{1}{2e \, m \, x}$ for all $x \in \mathbb{R}$, where $e$ is Euler's number. In the second inequality we have used $\frac{1}{d_\ell} = \frac{\widehat{\mu}_\ell +\lambda}{\widehat{\mu}_\ell} \leq \frac{1 + \lambda}{\widehat{\mu}_\ell}$, as $0 < \widehat{\mu}_\ell \leq 1$ for all $l = 1, 2,\ldots,N$.
\newline
\textbf{Convergence of $\rm II$:} 
\begin{align*}
&\mathbb{P}\left(\frac{1}{N}||U \left( I - \left(I-D\right)^m\right) U^\top \varepsilon^N ||^2_2 > \xi \right)
\\
&\leq \mathbb{P}\biggl(\left(\frac{1}{N}||U \left( I - \left(I-D\right)^m\right) U^\top \varepsilon^N ||^2_2 > \xi\right) \cap
\left\{ \mathbf{\widetilde{X}}^N \in \mathcal{B}_N \right\}\biggr)
+ \mathbb{P}\left(\left\{\mathbf{\widetilde{X}}^N \notin \mathcal{B}_N \right\} \right)
\end{align*}
The latter term goes again to $0$ by Assumption~\ref{ass:decay_eigenvalues}.
For any $\mathbf{\widetilde{x}}^N \in \mathcal{B}_N$ we show that 
$$
\mathbb{P}\left(\frac{1}{N} ||U \left( I - \left(I-D\right)^m\right) U^\top \varepsilon^N ||^2_2 > \xi\mid  \mathbf{\widetilde{X}}^N = \mathbf{\widetilde{x}}^N\right) \rightarrow 0
$$
for any $\xi > 0$ and uniformly in $\mathbf{\widetilde{x}}^N$. Recall that $\varepsilon^N \mid  \mathbf{\widetilde{X}}^N = \mathbf{\widetilde{x}}^N$ is a sub-Gaussian vector with mean $0$ and independent components. Thus, we can apply the Hanson-Wright inequality of Theorem~\ref{theorem:hanson_wright}. Remember that $D$ is a function of $\mathbf{\widetilde{x}}^N$. 
We need to show for any $\mathbf{\widetilde{x}}^N \in \mathcal{B}_N$, $A_m = \frac{1}{\sqrt{N}} U \left( I - \left(I-D\right)^m\right) U^\top $ that $ \mathbb{E}\left[\left\lvert\left\lvert A_m \varepsilon^N \right\rvert \right\rvert_2^2 \mid  \mathbf{\widetilde{X}^N} = \mathbf{\widetilde{x}}^N\right]  \rightarrow 0$, $||A_m^2||_F^2 \rightarrow 0$ and $||A_m^2|| \rightarrow 0$. 
It holds by Lemma~\ref{lemma:shrinkage} uniformly for $\mathbf{\widetilde{x}}^N \in \mathcal{B}_N$
\begin{align*}
\mathbb{E}\left[\left\lvert\left\lvert A_m \varepsilon^N \right\rvert \right\rvert_2^2 | \mathbf{\widetilde{X}} = \mathbf{\widetilde{x}}^N\right] &= \mathbb{E}_{Y^N\mid \mathbf{\widetilde{X}}^N = \mathbf{\widetilde{x}}^N}\left[ \left\lvert\left\lvert\frac{1}{\sqrt{N}} U \left( I - \left(I-D\right)^m\right) U^\top \varepsilon^N \right\rvert \right\rvert_2^2  \right] 
 \\
 &= \mathbb{E}_{Y^N\mid \mathbf{\widetilde{X}}^N = \mathbf{\widetilde{x}}^N}\left[ \frac{1}{N}tr\left(\left(\varepsilon^N\right)^\top U \left( I - \left(I-D\right)^m\right)^2 U^\top \varepsilon^N \right)  \right] \\
 &\leq \mathbb{E}_{Y^N\mid \mathbf{\widetilde{X}}^N = \mathbf{\widetilde{x}}^N}\left[ \frac{1}{N}tr\left(\varepsilon^N \left(\varepsilon^N\right)^\top\right) tr\left(U \left( I - \left(I-D\right)^m\right)^2 U^\top  \right)  \right] \\
&\leq  \frac{\sigma^2}{N}tr\left(\left( I - \left(I-D\right)^{m}\right)^2\right) \\
&= \frac{\sigma^2}{N}\sum_{\ell = 1}^N \left(1 - (1 - d_\ell)^m\right)^2 \\
&= \frac{\sigma^2}{ N}\sum_{\ell = 1}^N \left(1 - \left(1 - \frac{\widehat{\mu}_\ell}{{\lambda + \widehat{\mu}_\ell}}\right)^m\right)^2 \rightarrow 0
\end{align*}
for $N \rightarrow \infty$.
Further, 
\begin{align*}
||A_m^2||_F^2 &= \frac{1}{N^2} tr\left(\left( I - \left(I-D\right)^m\right)^4 \right)\leq \left(\frac{1}{N}tr\left(\left( I - \left(I-D\right)^m\right)^2 \right)\right)^2,
\end{align*}
which goes to $0$ with the same calculation as above. Finally, 
$$||A_m^2|| \leq \frac{1}{N} \rightarrow 0.$$ The statement follows by integrating out $\mathcal{B}_N$ with respect to $\mathbf{\widetilde{x}}^N$.

\end{proof}

%% file: Y_Appendix/D_algorithm.tex
\begin{algorithm}[H]
\DontPrintSemicolon
\label{algo:dagboost_large_p}
    \KwData{$\mathbf{x}^N = (\mathbf{x}_1, \ldots, \mathbf{x}_N)^\top \in \mathbb{R}^{N \times p}$}
    \KwIn{Kernels $K_1, \ldots, K_p$ implying RKHSs $H_1, \ldots, H_p$, penalty $\lambda$, step size $\mu$}
    \KwOut{$\widehat{G}, \widehat{F}$}
    $F^{(1)} \gets 0$\;
    $\widehat{G} \gets \emptyset$ \;
    $N \gets \emptyset$ \;
    $\widehat{f}_{kj} = \argmin_{g_{kj} \in H_j} \sum_{\ell = 1}^N \left(g_{kj}(\mathbf{x}_{\ell j}) - \mathbf{x}_{\ell k}\right)^2 + \lambda||g_{kj}||^2_{H_j}, j,k=1, \ldots, p, j\neq k$ \;
    $S(j, k) \gets \log\left(\sum_{\ell = 1}^N \left(\widehat{f}_{kj}(\mathbf{x}_{\ell j}) - \mathbf{x}_{\ell k} \right)^2 \right) $\;
    
    \For{$m\leftarrow 2$ \KwTo $m_{stop}$}{
    // Find the next edge and update graph\;
    $(j^0,k^0) \gets \argmin_{(j,k) \notin N} S(j,k, F^{(m-1)})$;
    $\widehat{G} \gets \widehat{G} + (j^0,k^0)$ \;
    $f^{(m)}_{k^0} \gets f^{(m-1)}_{k^0} + \mu \widehat{f}_{k^0j^0}$\;
    $f^{(m)}_{k} \gets f^{{(m-1)}}_{k}, k = 1, \ldots, k^0-1, k^0+1, \ldots, p$ \;
    $F^{(m)} \gets \left(f^{(m)}_{1}, \ldots, f^{(m)}_{p}\right)$\;
    // Check if AIC score has increased \;
    \lIf{$AIC(F^{(m)}, \mathbf{x}^N) > AIC(F^{(m-1)}, \mathbf{x}^N)$}
    {break}
    // Update edges that cause cycle and ensure they are not chosen anymore\;
    $N \gets getForbiddenEdges(\widehat{G}) $ \;
    // Update $S$\;
    $\widehat{f}_{k^0j} \gets$ Equation~\eqref{eq:candidate_functions_large_p}, $j = 1, \ldots, p$ \;
    $S(j, k^0, F^{(m)}) \gets $ Equation~\ref{eq:score_dagboost_large_p}, $j=1, \ldots, p$  \;
    }
    \KwRet{$\widehat{G}, F^{(m)}$}
\end{algorithm}